\pdfoutput=1
\documentclass[10pt,twocolumn]{IEEEtran}

\usepackage{cite}
\usepackage{amsmath,amssymb,amsfonts}
\usepackage{graphicx}
\usepackage{textcomp}
\usepackage{algorithmic}
\usepackage{array}
\usepackage{tabularx}
\usepackage{multirow}
\usepackage{booktabs}
\usepackage{hhline}
\usepackage{bm}
\usepackage{url}
\usepackage{stfloats}
\usepackage{placeins}

\usepackage{capt-of}

\usepackage{cuted}
\usepackage{subcaption}

\graphicspath{{figures/}}

\newcolumntype{Y}{>{\raggedright\arraybackslash}X}
\newcolumntype{P}[1]{>{\raggedright\arraybackslash}p{#1}}
\newcolumntype{C}[1]{>{\centering\arraybackslash}p{#1}}

\usepackage[font=footnotesize,labelfont=bf,labelsep=period]{caption}

\title{Pix2Pix-Hybrid: Structure-Guided Conditional Synthesis of Hajj Crowd Images with Multi-Channel Conditioning and Weak Attribute Supervision}

\author{
\begin{minipage}{0.98\textwidth}
\centering
Amirah F. Alshammari$^{1,2}$, Bander A. Alzahrani$^{3}$, and Nahed A. Alowidi$^{4}$\\[0.8em]
\small
$^{1}$Information Systems Department, Faculty of Computing and Information Technology, King Abdulaziz University, Jeddah, Saudi Arabia\\
$^{2}$Department of Information Systems, College of Computer and Information Sciences, Jouf University, Aljouf, Saudi Arabia\\
$^{3}$Information Systems Department, Faculty of Computing and Information Technology, King Abdulaziz University, Jeddah, Saudi Arabia\\
$^{4}$Computer Science Department, Faculty of Computing and Information Technology, King Abdulaziz University, Jeddah, Saudi Arabia\\[0.5em]
E-mails: aalshammari0114@stu.kau.edu.sa, baalzahrani@kau.edu.sa, nalowidi@kau.edu.sa
\end{minipage}
}

\date{}

\begin{document}

\renewcommand{\thetable}{\arabic{table}}
\renewcommand{\thefigure}{\arabic{figure}}

\maketitle

\begin{abstract}
Developing accurate crowd-counting models for Hajj pilgrimage scenes remains challenging due to the lack of domain-specific annotated images. In addition, manual collection of data during large gatherings is difficult and raises privacy concerns related to the identification of individuals. To address these limitations, this paper proposes Pix2Pix-Hybrid (P2P-H), a hybrid conditional GAN for structure-guided Hajj crowd image synthesis for training and data-augmentation purposes. Our model builds on the Pix2Pix architecture and incorporates several improvements in both structure and training. P2P-H employs a U-Net generator conditioned on eight input channels that jointly encode structural cues (edges and grayscale) and contextual attributes (crowd density and time of day). To better capture highly detailed pattern textures in extremely dense scenes our approach integrates two multi-scale PatchGAN discriminators operating at different resolutions. The training procedure combines adversarial, perceptual, and feature-matching objectives within a hybrid optimization scheme that incorporates adaptive data augmentation and stabilization strategies. The model was trained on 993 real Hajj frames collected from 60 publicly available video sources with conditioning attributes derived automatically to reduce manual labeling effort. Using this framework, we constructed CrowdH, a synthetic dataset of 10,000 high-resolution Hajj crowd images. The experimental results show that P2P-H improves structure-preserving conditional synthesis quality compared with the evaluated Pix2Pix and StyleGAN2-ADA baselines, while showing favorable transfer to other crowd datasets. Moreover, to assess practical utility beyond image-level generation quality, we further constructed CrowdH-Mix-469, an annotated mixed real-synthetic dataset comprising 384 real Hajj images and 85 selected synthetic images and evaluated five representative crowd-counting models under two training settings: real-only and real-plus-synthetic. Experimental results show that the proposed synthetic data improve downstream counting performance in terms of MAE across all five models, with the strongest gain observed for CSRNet. Overall, the proposed framework provides a scalable, privacy-aware, and structure-guided conditional approach for Hajj crowd-data augmentation with practical support for downstream crowd-counting model training.
\end{abstract}

\vspace{0.2em}

\section{Introduction}
Accurate crowd counting is essential for public safety, infrastructure management, and operational decision-making in large-scale events.  This need is particularly critical during the Hajj pilgrimage, where millions of individuals gather in a limited area.  Effective monitoring of crowd density and flow helps to prevent overcrowding hazards, support emergency response and ensure that transportation and facility management systems operate safely and efficiently\cite{b1}. Despite the success of deep learning (DL)\cite{b2} in vision-based crowd analysis, the reliability of existing models is constrained by the scarcity of domain-specific annotated datasets, privacy concerns, and the significant cost and time associated with manual labeling. Furthermore, Hajj-specific crowd images exhibit extreme crowd densities, heavy occlusion, and illumination variability, which limit the generalizability of models trained on open-source benchmarks, such as ShanghaiTech or UCF-QNRF \cite{b3},\cite{b4}.

A common strategy for data scarcity is data augmentation techniques; however, augmented samples remain strongly correlated with the original data and fail to introduce new structural or contextual variations \cite{b5}. As a result, the diversity required to model complex crowd scenes is lacking. Synthetic data generation offers an alternative strategy for expanding limited training dataset. Generative models learn the statistical distribution of real images to produce realistic synthetic samples \cite{b6}. Two main types of generative modeling methods have become popular for image synthesis: Generative Adversarial Networks (GANs) \cite{b8} and diffusion models \cite{b10}. Both methods can produce high-quality, photorealistic results; however, each method has limitations. Diffusion models, despite their strong fidelity and convergence properties, have been shown to overfit training data, particularly in low-data regimes \cite{b11}, \cite{b12}. However, GANs often suffer from mode collapse, resulting in less variability in the synthesized outputs because the generator cannot fully capture the diversity of the target distribution \cite{b7}, \cite{b9}. Moreover, because both approaches require the training of deep neural architectures, their effectiveness is strongly dependent on the size and quality of the available training data. Limited or imbalanced datasets can degrade generative performance, introduce distributional biases, and reduce the generalizability of downstream models.

Among generative architectures, conditional GANs (cGANs) have demonstrated strong performance in image-to-image translation tasks in which structured inputs such as an edge map or grayscale layout are mapped onto a corresponding photorealistic output. The P2P framework introduced this approach by training a U-Net generator and PatchGAN discriminator under adversarial and reconstruction objectives, enabling direct pixel-level translation between paired data \cite{b7}. However, despite its effectiveness in simple settings, P2P often produces blurred textures and checkerboard artifacts when trained on complex, high-resolution images such as crowd scenes.

To overcome these limitations, P2PHD \cite{b13} applied multi-scale PatchGAN discriminators. The discriminators simultaneously analyze realism at multiple resolutions for each image. This allows P2PHD to achieve both high-level global consistency and local detail, thereby making the training process much more stable. Furthermore, the P2PHD architecture offers the ability to effectively generate high-resolution images. In addition to P2PHD enhancements, SGAN2-ADA \cite{b9} achieves high-fidelity performance with minimal training data. SGAN2-ADA achieves this by applying adaptive discriminator augmentation (ADA). ADA automatically modifies the strength of augmentation during training to minimize overfitting caused by the limited number of samples available for training. As a result, the model generalizes well, even with limited training data. Enhanced normalization and regularization are also included in the architecture to reduce the likelihood of mode collapse \cite{b9}; therefore, the quality of the output remained consistent across all domains. Due to these characteristics, the model is suitable for limited-data settings, such as Hajj crowd imagery, where data availability is constrained by privacy restrictions.

In contrast to fully unconstrained image generation, the objective of this study is structure-preserving conditional synthesis for Hajj crowd-data augmentation. The proposed framework does not aim to generate arbitrary crowd layouts from random latent vectors. Instead, it renders Hajj crowd appearances from structural and weak semantic conditions derived from real crowd images, with the goal of improving data availability and downstream crowd-counting training under privacy-sensitive conditions.

Building on Pix2Pix(P2P),Pix2PixHD(P2PHD), and ADA-style adversarial stabilization, we develop P2P-H as a domain-specific structure-guided conditional synthesis framework for Hajj crowd-image augmentation. The proposed framework is not intended to be a fully unconstrained latent generative model. Instead, it focuses on rendering realistic Hajj crowd appearances from self-derived structural and weak semantic conditions, while preserving the dominant scene geometry required for downstream crowd-counting tasks.

Our framework integrates three complementary ideas within a unified architecture: (i) a P2P-style conditional backbone with adversarial and reconstruction losses. In our setting, the supervision pairs are self-paired. The conditioning tensor is derived from the same target image (x = g(y)). Unlike traditional paired image-to-image translation methods that require independently captured image pairs (e.g., day/night photographs of the same scene), our approach performs self-paired conditional reconstruction, where both the edge-based input and RGB target are derived from the same source image during training. This self-pairing strategy enables training without requiring expensive paired datasets while maintaining conditional guidance over the generation process. In this framework, the structural configuration of the crowd is intentionally preserved through the conditioning channels, while the generator learns to produce appearance-diverse crowd renderings under different contextual attributes such as density category and time-of-day.  (ii) the multi-scale adversarial training strategy of P2PHD, which employs discriminators operating at different resolutions to enforce both global consistency and local realism; and (iii) adaptive regularization and augmentation mechanisms inspired by StyleGAN2-ADA (SGAN2-ADA), which stabilize training and improve generalization when only a limited number of real images are available. The generator is conditioned on multi-channel structural and contextual inputs that are automatically extracted. This conditioning enables P2P-H to produce structurally coherent crowd images with attribute-driven appearance variation while reducing reliance on manually labeled attributes.

This study makes several contributions to the application of generative modeling and synthetic dataset development for crowd analysis in challenging, privacy-sensitive environments, which can be outlined as follows:

\begin{itemize}

\item {We present a domain-specific, structure-guided conditional synthesis framework based on a hybrid adversarial design that integrates established components, including self-paired conditional supervision, multi-channel structural-context conditioning, multi-scale adversarial discrimination, adaptive regularization, perceptual and feature-matching objectives, and EMA-based stabilization. These components are adapted for limited-data Hajj crowd augmentation in a challenging, privacy-sensitive domain, and the contribution lies in the task-specific design, integration, and empirical validation of the framework.}

\item {We developed an automatic weak-supervision pipeline that extracts structural guidance (edge and grayscale representations) and contextual attributes representing crowd density (low/medium/high) and time-of-day information (morning/afternoon/night) from unlabeled images, thereby reducing manual annotation effort.}
\item {We designed a multi-level evaluation protocol to assess conditional fidelity, distributional similarity, appearance-level variation, semantic controllability, cross-domain behavior, memorization risk, downstream crowd-counting utility.The protocol also includes supporting interpretability analyses, such as real–fake classification and feature-space visualization using t-SNE.}
\item {We constructed CrowdH, a synthetic dataset of 10,000 high-resolution Hajj crowd images generated by the proposed P2P-H framework under structure-guided conditioning to support data augmentation and future crowd-analysis research.}
\item {We further constructed CrowdH-Mix-469, an annotated mixed real-synthetic evaluation set, and used it to assess the practical utility of the generated images for crowd-counting.}
\end{itemize}
The remainder of this paper is organized as follows. Section II reviews related work on crowd counting datasets and GAN-based image synthesis. Section III describes the proposed P2P-H architecture, training methodology, and evaluation protocol. Section IV presents the experimental results and discusses the findings, including quantitative metrics, qualitative visualizations, and comparative analysis. Section V outlines the limitations of the proposed framework. Finally, Section VI concludes the study with key insights and directions for future research.

\section{Related Work}

The development of robust crowd counting and density estimation systems depends on high-quality datasets that capture diverse crowd scenarios. In addition, efficient annotation strategies reduce manual labeling costs. This section systematically reviews existing work across these dimensions, highlighting its contributions and limitations in the context of crowd analysis.

\subsection{EXISTING DATASETS FOR CROWD COUNTING AND HAJJ SCENE ANALYSIS}
Real Crowd Counting Datasets: Prior work on crowd counting and density estimation has advanced rapidly with the availability of benchmark datasets and deep architectures. This progress remains tightly coupled with how representative and well-annotated the training data are. Early crowd counting datasets such as UCSD \cite{b15} , Mall \cite{b16}, and UCF\_CC\_50 \cite{b14} have served as important benchmarks, but they are limited by low crowd densities, restricted camera viewpoints, or a small number of images. The WorldExpo’10 data set \cite{b17} increased scene diversity through multi-camera coverage, but it was still limited to moderate crowd densities and specific event environments. More recent datasets, including ShanghaiTech Part B \cite{b3}, UCF-QNRF \cite{b4}, JHU-CROWD++ \cite{b18}, and NWPU-Crowd \cite{b19}, have expanded both the number of images and the level of detailed annotation, supporting the growth of deep learning methods in this field. However, these datasets mainly feature crowds in urban or public settings, where people wear various clothing and move randomly. Consequently, models trained on them often fail when applied to Hajj footage because of differences in appearance, occlusions, and spatial layout.

Hajj-Specific Dataset: Hajj-focused datasets attempt to close this gap but remain limited in scope, access, or annotation depth. The HUER dataset provided event-level annotations for multiple Hajj pilgrimage rites and human actions, but its limited scale and restricted access reduce its usefulness for comparative research \cite{b20}.  The HAJJ-Crowd dataset introduced real surveillance images with head-bounding boxes and density maps for CNN-based crowd density estimation in the Tawaf area. However, this dataset is limited in scope, small in scale, and not publicly available \cite{b21}. Further studies used data augmentation and manual labeling to expand the dataset size, but these approaches sometimes introduced artificial patterns and relied on subjective classifications. Additionally, they often fail to cover other important Hajj locations such as Sa'i, Arafat, Mina, and Muzdalifah \cite{b22}, \cite{b23}. A recent study also introduced multi-modal and behavioral datasets, but important challenges remain. Other Hajj-related resources include video-centric sets (e.g., HAJJv1/v2) that emphasize behavioral cues and optical flow, but camera viewpoints are fixed and detailed density annotations are scarce \cite{b24}, \cite{b26}. The wearable sensor dataset captures physiological signals from ritual participants; however, short recording sessions and a narrow participant pool affect the representativeness of the data \cite{b26}. The Hajj and Umrah Crowd Management dataset expanded the scale to 10,000 images with head-point annotations \cite{b27}, but issues with inconsistent labeling and the absence of standardized density maps have remained challenging for precise training and evaluation. The existing Hajj datasets are limited in size, accessibility, scene diversity, and annotation quality. This highlights the essential need for a large, well-annotated, and contextually relevant Hajj crowd image dataset to support deep-learning research on crowd counting and safety management during the pilgrimage.

Synthetic Crowd Counting Datasets: Generating data offers an alternative way to reduce annotation costs and expand data diversity. Several simulation-based synthetic datasets have been developed for crowd counting. Simulation-based methods use game engines to create synthetic crowd scenes. These methods provide synthetic images with automatic labels and controllable variations in crowd density, viewpoint, and lighting, such as GCC \cite{b28}, CrowdX \cite{b29}, and CVCS \cite{b30} datasets. While these datasets provide large, automatically labeled training sets, they are primarily urban, game-engine environments and exhibit domain gaps with pilgrimage scenes. Therefore, none of these synthetic datasets provide large, high-quality visual representations matching the complexity of the visual characteristics (density, appearance, and spatial organization) found in Hajj-like environments. This has increased the need to develop new synthetic data generation methods that better reflect the characteristics of the target environment.

Table~\ref{tab:datasets} summarizes representative existing datasets. Real, Hajj-specific, and synthetic datasets have advanced crowd counting research. Both real-world and synthetic datasets primarily focus on urban crowds. The Hajj-specific datasets are smaller in scale or remain inaccessible. Across all categories, there is an absence of a dataset that combines scale, diversity, and Hajj authenticity. This gap motivates our work: we aim to synthesize high-fidelity Hajj crowd images, to support safe crowd analysis during the pilgrimage.

\subsection{GENERATIVE ADVERSARIAL NETWORKS (GANs) FOR IMAGE SYNTHESIS }
Generative Adversarial Networks (GANs) have become standard tools for realistic image generation. The GAN framework was first introduced in \cite{b6}, which established a two-part architecture comprising a generator and a discriminator trained adversarially. The generator learns to produce data that mimics the distribution of real samples, while the discriminator aims to distinguish between real and synthetic data, thereby improving the performance of the generator. Conditional GANs (cGANs) \cite{b37} extend this idea by incorporating additional contextual information into the model and allowing the creation of synthetic images with a target class or attribute. Further advancements in GANs include DCGANs \cite{b38}, which utilize convolutional neural networks to generate more stable and realistic images. Another advancement is CycleGANs \cite{b39}, which support unpaired image-to-image translation, and therefore can be applied to tasks such as style transfer and domain adaptation, when paired data are unavailable.

The P2P framework \cite{b7} is another popular approach to paired translation tasks. The authors mapped structured inputs to realistic outputs using both adversarial and L1 reconstruction objectives. P2P has been widely applied in many fields, including medical imaging \cite{b40}, \cite{b41}, agricultural analysis \cite{b42}, and crowd density estimation \cite{b31}. However, standard P2P may produce blurry textures and checkerboard artifacts in complex scenes because of the limitations of using only a single-scale discriminator and structural conditioning. To address the drawbacks of the P2P framework, P2PHD \cite{b13} introduced multi-scale PatchGAN discriminators to assess realism at both global and local scales, resulting in sharper textures and improved spatial coherence. Additionally, recent advances such as SGAN2-ADA \cite{b9}, has also improved the training stability under limited data by utilizing adaptive discriminator augmentation, reducing overfitting, and improving generalization with smaller datasets. These recent developments in GANs have contributed to a transition toward synthesis, which is both data-efficient and controllable for desired attributes.

\begin{table*}[!t]
\caption{Summary of representative real and synthetic datasets for crowd counting.}
\label{tab:datasets}
\centering

\scriptsize
\setlength{\tabcolsep}{2.2pt}
\renewcommand{\arraystretch}{1.18}
\setlength{\arrayrulewidth}{0.4pt}

\begin{tabular}{|
P{0.135\textwidth}|
P{0.235\textwidth}|
P{0.115\textwidth}|
C{0.045\textwidth}|
P{0.255\textwidth}|
C{0.075\textwidth}|}
\hline
\textbf{Dataset} & \textbf{Images/Frames/Videos} & \textbf{Type} & \textbf{Hajj} & \textbf{Objective} & \textbf{Public} \\
\hline
UCSD \cite{b15} & 2000 frames (49,885 pedestrians) & Real & No & Crowd counting & Yes \\
\hline
PETS \cite{b32} & 8 videos & Real & No & Crowd counting & Yes \\
\hline
UCF \cite{b4} & Multiple videos & Real & No & Crowd counting, activity recognition, anomaly detection, motion prediction & Yes \\
\hline
UCF\_CC\_50 \cite{b14} & 50 images & Real & No & Crowd counting & Yes \\
\hline
Mall \cite{b16} & 2000 frames & Real & No & Crowd counting & Yes \\
\hline
WWW \cite{b33} & 8,257 scenes, 8M frames (10,000 videos) & Real & No & Crowd counting, crowd motion analysis & Yes \\
\hline
ShanghaiTech \cite{b3} & Part A: 482, Part B: 716 images & Real & No & Crowd counting & Yes \\
\hline
WorldExpo'10 \cite{b17} & 245 scenes, 2630 videos & Real & No & Crowd counting & Yes \\
\hline
Crowd-11 \cite{b34} & 3005 scenes, 6272 videos, 11 classes & Real & No & Crowd counting & Yes \\
\hline
JHU-CROWD++ \cite{b18} & 4372 images & Real & No & Crowd counting & Yes \\
\hline
NWPU-Crowd \cite{b19} & 5109 images & Real & No & Crowd counting & Yes \\
\hline
GCC \cite{b28} & 15,212 images (7.6M persons) & Synthetic & No & Pretraining, domain adaptation & Yes \\
\hline
CrowdX \cite{b29} & 30,000 images (parametric variation) & Synthetic & No & Factor impact analysis & Upon request \\
\hline
CVCS \cite{b30} & 31 scenes $\times$ 100 views $\approx$ 310,000 frames & Synthetic (multi-view) & No & Cross-scene and cross-view density fusion & Yes \\
\hline
CrowdSim2 \cite{b35} & Thousands of frames (5 weather scenarios) & Synthetic & No & Object detection under varied densities & Yes \\
\hline
TUB CrowdFlow \cite{b36} & 3200 frames (10 UAV sequences) & Synthetic (video) & No & Optical flow, trajectory ground truth & Yes \\
\hline
HUER \cite{b20} & Images and videos from six rituals & Real Hajj data & Yes & Activity recognition, anomaly detection & No \\
\hline
HAJJ-Crowd \cite{b21} & 1500 images, 10 video sequences & Real Hajj data & Yes & Density estimation, benchmarking & No \\
\hline
HAJJv1 \cite{b24} & Not stated & Real Hajj data & Yes & Abnormal behavior detection & Yes \\
\hline
HAJJ-Crowd \cite{b22} & 27,000 images (21,600 train; 5400 test), 25 videos & Real / Augmented Hajj data & Yes & Density analysis & No \\
\hline
HAJJv2 \cite{b25} & 18 videos (170k train, 130k test annotations) & Real Hajj data & Yes & Track behavior, classification & Yes (CSV) \\
\hline
Hajj Activity Prediction \cite{b26} & 64 participants, multimodal sensors (10 s windows) & Real Hajj data & Yes & Activity classification & Yes (CSV) \\
\hline
\end{tabular}
\end{table*}
\FloatBarrier


GANs have increasingly been used in crowd-counting applications to augment data and model complex spatial temporal patterns. Recent augmentation studies applied P2P GANs to translate RGB crowd images into fake thermal images and improved the low light counting accuracy by approximately 5\% \cite{b31}. However, these fake thermal images still do not match the real noise and thermal patterns of the actual sensor data, especially when paired thermal data are unavailable. In another study, a P2P GAN was applied to denoise hazy or noisy crowd images before counting. In addition, P2P was also used in \cite{b57} to translate RGB crowd scenes into synthetic thermal images and feeding the GAN-enhanced images into standard density estimators yielded improved counting results. Similarly,\cite{b58} developed a GAN-based optical flow framework (using U-Net/FlowNet generators) to simulate and distinguish between normal and abnormal motions in very large Hajj pilgrim crowds. More recently, Fusion Counting was introduced in \cite{b59} as a multi-task RGB–thermal fusion network that incorporates adversarial loss to jointly improve image fusion and density estimation, yielding more robust counting under challenging conditions. These studies underscore the potential of adversarial methods for crowd datasets and and enhancing counting performance. However, most current GAN-based approaches still depend on extensive manual labels (for example, paired RGB-thermal images or flow annotations) to guide the conditional generation process.

Recent image-synthesis research has been increasingly influenced by diffusion-based models, which have demonstrated strong performance in high-fidelity and conditional generation \cite{b10}, \cite{b11}. A recent comparative study of domain-specific scientific image synthesis further indicates that the choice of generative architecture involves task-dependent trade-offs: GAN-based models can provide high perceptual quality and structural coherence, whereas diffusion-based methods can achieve high realism but may require careful domain-specific validation to ensure that visually plausible outputs remain faithful to the target domain \cite{b69}. In crowd-analysis research, diffusion-based approaches have also begun to appear. CrowdDiff formulates crowd density estimation as a denoising diffusion process and demonstrates the relevance of diffusion models to density-map-based crowd counting \cite{b63}. In a related direction, Wang et al.\cite{b64} used a ControlNet-based diffusion framework conditioned on head-location dot maps to generate synthetic images for crowd-counting augmentation and reported downstream improvements on several benchmark datasets. These studies indicate that diffusion models represent an important direction for modern crowd-image synthesis and counting-oriented data augmentation. 

However, the objective of the present work differs from open-ended image generation. P2P-H addresses structure-guided conditional synthesis under self-paired inputs for dense Hajj scenes, where preserving the dominant crowd layout is important for downstream crowd-counting analysis. In this setting, the edge and grayscale channels are not auxiliary prompts, but primary structural constraints used to preserve scene geometry, and the task is formulated as a direct paired mapping x→y from a fixed structural conditioning tensor to an RGB crowd image. This formulation is naturally aligned with the conditional image-to-image translation paradigm for which the P2P/P2PHD family was designed \cite{b7},\cite{b13}, because it directly supports adversarial, reconstruction, perceptual, and feature-matching objectives for structure-preserving rendering. In addition, ADA-style adversarial stabilization has been shown to be useful in limited-data regimes \cite{b9}, which supports the use of an adversarial formulation under the limited, single-domain Hajj dataset considered in this study.
We do not claim that GAN-based methods are generally superior to diffusion-based synthesis. Rather, the use of a GAN-based formulation reflects task alignment with the structure-preserving conditional synthesis objective and the limited-data conditions considered in this study. A dedicated, task-matched comparison with fine-tuned diffusion and ControlNet-style baselines under the same conditioning, resolution, training data, and downstream-counting protocol remains an important direction for future work.

\subsection{WEAK SUPERVISION AND AUTOMATED ATTRIBUTE LEARNING}
Despite these improvements, most conditional GANs \cite{b37}  rely on extensive manual labeling for attribute-controlled generation. In crowd analysis, particularly within the Hajj context, obtaining accurate annotations for density or illumination is infeasible due to occlusions and subjectivity in human labeling. To address this challenge, recent research has leveraged weak supervision and self-supervised learning to automatically infer attributes from data. For instance, \cite{b43}, \cite{b44} demonstrated that conditional generation can be used to effectively guide weak or paired labels data, while \cite{b45} investigated the unsupervised discovery of interpretable latent directions in GANs. Gao et al. \cite{b60} introduced a domain-adaptive crowd counting approach that uses two-step GAN-driven image translation to synthesize high-fidelity crowd scenes from simulations and then reconstruct density maps. This enables robust counting in unseen domains with minimal new annotations.

In Hajj crowd-counting research, the learning abilities of GANs have not yet been adopted. The primary objective of this research is to use GANs to increase the number of datasets and improve model training. This objective was prioritized due to the remarkable capability of GANs to estimate complex probability distributions of visual data.  Therefore,the proposed framework follows a structure-guided conditional generation paradigm, in which the model learns to translate structural representations into realistic crowd images. The model automatically extracts eight conditioning channels comprising structural features and contextual attributes. These channels guide the generator to produce realistic crowd images while preserving crowd structure.

\section{Methodology}
\begin{figure}[!htbp]
\centering
\includegraphics[width=\columnwidth]{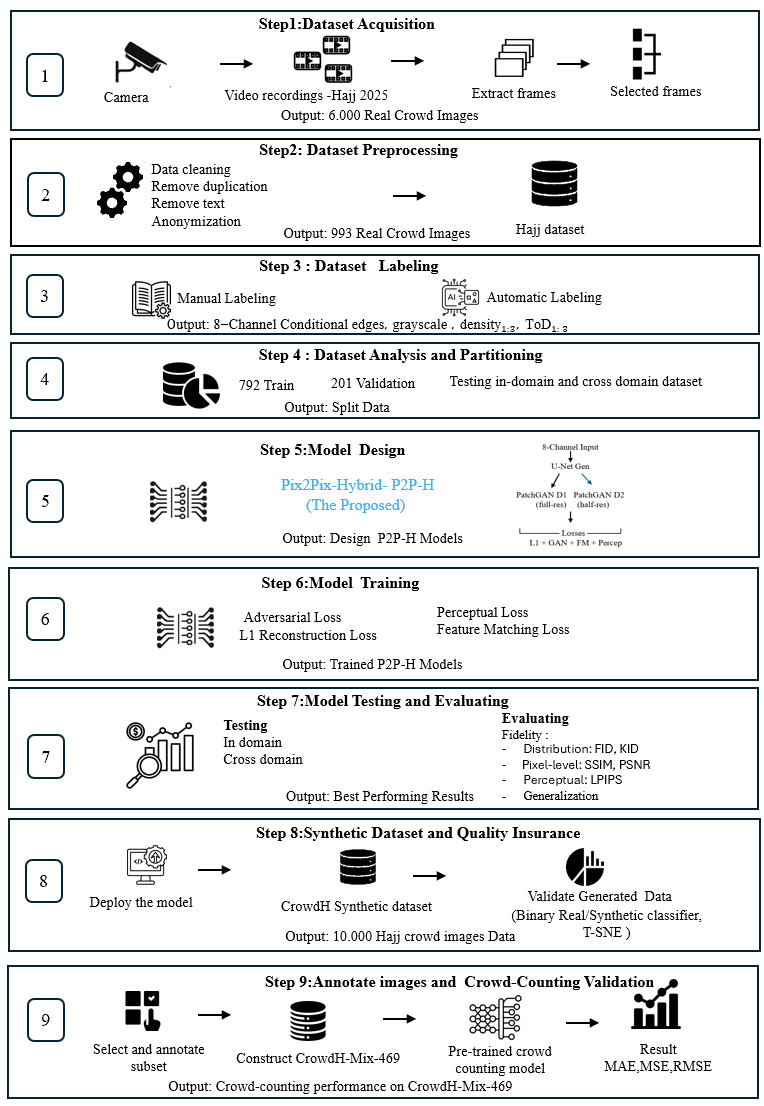}
\caption{Overview of the proposed end-to-end framework for Hajj crowd image generation, synthetic data validation, and downstream crowd-counting evaluation using GAN-based methods.}
\label{Fig_1}
\end{figure}

The methodology of this study aims to construct a high-fidelity, attribute-conditioned synthetic Hajj crowd dataset for crowd-counting research. The proposed pipeline is illustrated in Fig.~\ref{Fig_1}. It consists of nine steps. Step 1 involved collecting Hajj videos and extracting frames from them. Then, these frames were preprocessed through duplicate removal, text 
elimination, and anonymization, resulting in a curated set of 993 high-quality crowd images (Step 2). In Step 3, data labeling using both manual and automated techniques produced a multi-channel conditional input that represents crowd density (low/medium/high) and time-of-day information (morning/afternoon/night) along with edge and grayscale features. In Step 4, we divided the dataset into 792 images for training and 201 images for validation. Three open-source datasets were used for the in-domain and cross-domain testing.Model design and training were implemented in Steps 5 and 6. The proposed P2P-H architecture incorporated a U-Net generator and two PatchGAN discriminators. These components were optimized using a composite loss function that integrates adversarial, perceptual, and feature-matching objectives with adaptive weight scheduling. The trained model was then tested and evaluated in Step 7 using three benchmark datasets HAJJv2, UCF\_CC\_50, and UCF-QNRF. HAJJv2 was used to test domain-specific performance. UCF\_CC\_50 and UCF-QNRF were used to assess generalization.

Furthermore, multiple quality metrics were used to ensure fidelity, diversity, and generalization. Finally, the best model was deployed to produce the CrowdH synthetic dataset, which was validated through qualitative and quantitative assessments (Step 8).In Step 9, a crowd-counting validation was conducted by constructing CrowdH-Mix-469, an annotated mixed real-synthetic dataset. This dataset was then used to evaluate the practical utility of the generated images by applying  five representative crowd-counting models under two settings, namely real-only and real-plus-synthetic, with testing performed exclusively on the held-out real Hajj test set.

\FloatBarrier

\subsection{DATASET ACQUISITION}
Due to restricted access to direct camera feeds within the Hajj pilgrimage area, this study used publicly available video content. Sixty high-resolution YouTube videos containing crowd scenes from different locations were selected based on strict qualitative criteria.  All videos were then reviewed manually to determine whether they met the following visual quality criteria: (i) high resolution with low perceptible compression artifacts; (ii) stable wide-angle views; (iii) clearly visible crowd formations; (iv) spatial and temporal diversity across key Hajj locations (e.g., Arafat, Mina, Tawaf, Al-Jamarat, Safaa-Marwah); and (v) consistent lighting conditions. After the manual review process, frame extraction was completed using the VLC media player's Scene Filter at a sampling rate of one frame per thirty frames. Finally, a total of 6.000 raw frames were collected.

\subsection{DATASET PREPROCESSING }
Data Cleaning: The extracted frames require cleaning to ensure dataset quality. To achieve this goal, we performed four data-cleaning steps: removing similar images, deleting duplicate images, quality filtering, and removing text from images. First, frames that looked similar, especially those captured from static scenes or at short intervals, were manually reviewed and removed. Second, redundant samples were removed to minimize dataset duplication. Third, quality filters were applied to the images. Each frame with blurred or insufficient resolution was excluded based on visual assessment. Finally, we used a text cleanup tool \cite{b54} to remove embedded text such as timestamps, watermarks, and channel identifiers to avoid bias when training our model. The four-step cleaning process resulted in a reduction of 6,000 raw frames to 993 high-quality and diverse images for our dataset.

Privacy-Preserving Anonymization: To protect individual identities within crowds, a face-blur anonymization method was employed. First, frames with identifiable faces were manually selected. Next, an automated face blurring algorithm was employed to blur the faces \cite{b55}. This anonymization procedure removed identifiable facial information while preserving the spatial structure and density characteristics of the crowd scene. The objective was therefore to reduce identity-related privacy risks while maintaining the structural information necessary for crowd analysis tasks.

\subsection{DATASET LABELING AND SELF-PAIRED CONDITIONING CONSTRUCTION}
Automatic Weak Labeling: To overcome annotation challenges in large-scale Hajj imagery, an automatic weak-labeling pipeline was developed to extract two semantic attributes without manual supervision. It consists of two attribute extraction processes: (i) Crowd-Density Estimation: Each RGB image was first converted to grayscale, smoothed with a Gaussian filter ($\sigma$= 1.0, 5×5 kernel), and processed using Canny edge detection with thresholds 64/128. The edge-density ratio $\rho_e$ as given in (1) was then computed as the proportion of edge pixels to total image area, where E(x,y) denotes the binary edge map.

\noindent

\begin{equation}
\tag{1}\label{eq:rho_e}
\rho_{\mathrm{e}}
=
\frac{\left|\left\{(x,y)\,:\,E(x,y)>0\right\}\right|}{H \times W}.
\end{equation}

To obtain categorical crowd-density labels, percentile-based thresholds (33.33\% and 66.67\%) were used to adaptively partition the dataset into three discrete levels (low, medium, and high), ensuring balanced representation across different density levels. (ii) Temporal Illumination Inference: We computed the Y (luma) channel from YCrCb and computed its mean $\mu_{\mathrm{Y}}$; dataset-wide percentiles ($(\theta_1,\theta_2)$) were used to discretize $\mu_{\mathrm{Y}}$ into three classes: 0 represents morning/day (bright), 1 indicates afternoon (moderate), and 2 represents night/dim (low-light). Both attributes were one-hot encoded (three channels each), spatially broadcast to the image size, and concatenated with the Canny edge and grayscale channels to form an eight-channel conditioning tensor as given by (2). This multi-channel representation provides complementary structural (edges and gray) and contextual (density and time-of-day) guidance to the generator.

\begin{equation}
\mathbf{x}
=
\bigl[\mathbf{1}_{\mathrm{edge}}+\mathbf{1}_{\mathrm{gray}}
+\mathbf{3}_{\mathrm{density}}+\mathbf{3}_{\mathrm{time}}\bigr]
\in [-1,1]^{768\times768\times8}.
\label{eq:input_tensor}
\end{equation}

Self-Paired Input Construction: We employed a self-pairing strategy due to the absence of externally given aligned input–output pairs for large-scale Hajj scenes. Specifically, for each target image y, we deterministically construct its conditioning tensor x = g(y) by extracting (i) structural cues (Canny edges and grayscale) and (ii) weak semantic attributes (density and time-of-day) from the same image y. Therefore, training uses pairs (x, y), where x is derived directly from y and y is used as the supervision target. All channels are normalized to $[-1, 1]$ before training. In our work, the input condition is derived from the target image itself, and the model is trained to reconstruct that original image. Because the conditioning tensor is deterministically derived from the target during training,the learning objective primarily captures conditional reconstruction by inverting edge/gray preprocessing, and attribute variation is limited to the information encoded in the conditioning channels rather than free-form stochastic sampling. This differs from the common paired translation setting where x provides  an external conditioning modality (e.g., segmentation/label maps) paired with y \cite{b7}, rather than being deterministically derived from the target y. Unlike unconditional GANs \cite{b9}, which synthesize from a latent noise vector z, our generator receives no noise input and is fully conditioned on x.

Manual Validation: The 993 automatically labeled images were manually reviewed for quality assurance, with labels adjusted where automatic extraction failed due to extreme lighting or occlusion. All 993 images were manually organized by location, density level, and image index. The crowd density categories were defined as follows: Class 0 for low density, Class 1 for medium density, and Class 2 for high density. This approach provides structured supervision without computationally expensive precise counting.
\subsection{DATASET ANALYSIS AND PARTITIONING }
\begin{figure}[!htbp]
\centering
\includegraphics[width=\columnwidth,height=5cm]{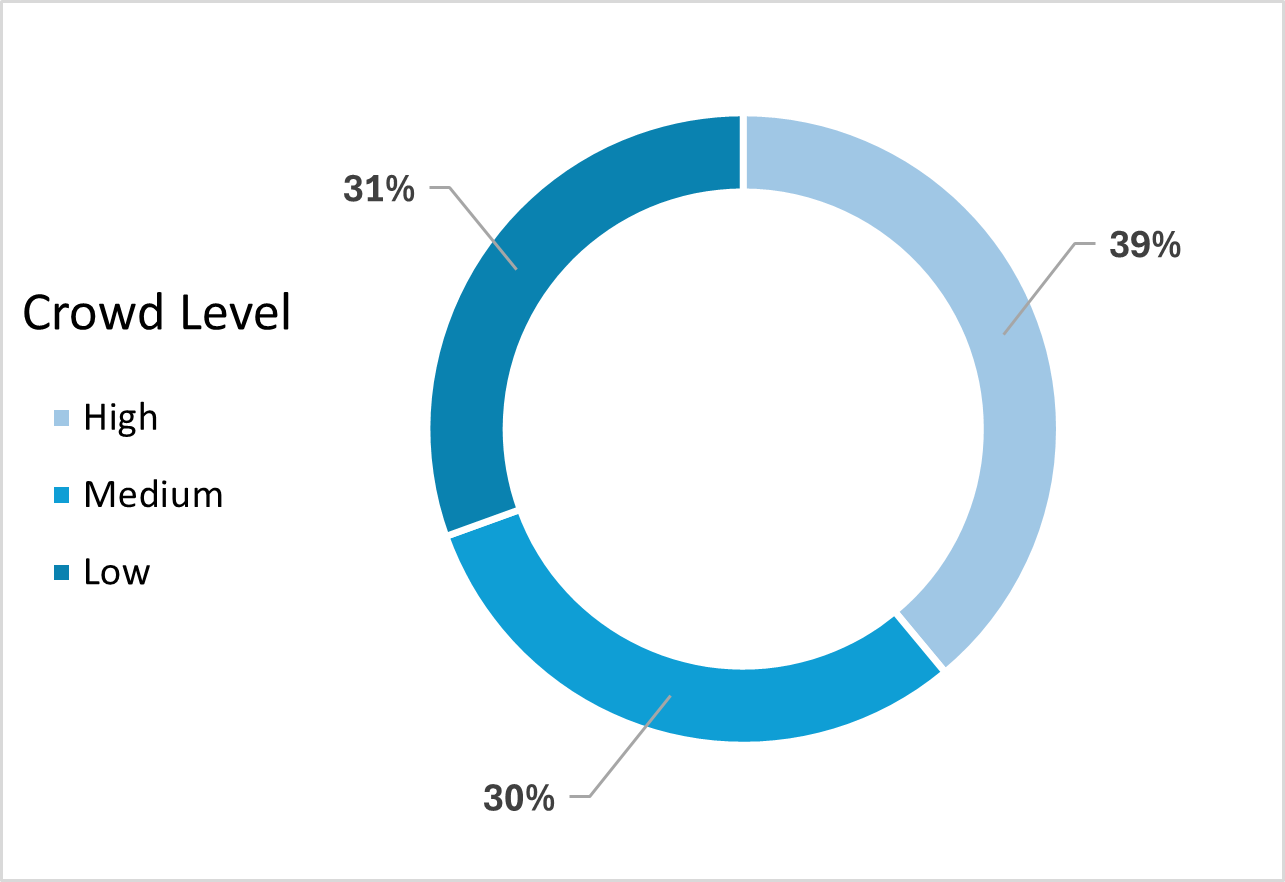}
\caption{Training Set Density Distribution.}
\label{fig:train_density}
\end{figure}

Dataset Analysis: Statistical analysis was applied to the data collection of 993 images to understand the dataset properties in terms of density and location coverage. The density of the training set is shown in Fig.~\ref{fig:train_density}, where the density distribution is uniform: 30.5\% are low-density scenes, 30.5\% are medium-density scenes and 39.4\% are high-density scenes. This balance across density levels supports robust learning across diverse crowd condition scenarios.

\begin{figure}[!htbp]
\centering
\includegraphics[width=\columnwidth,height=5cm]{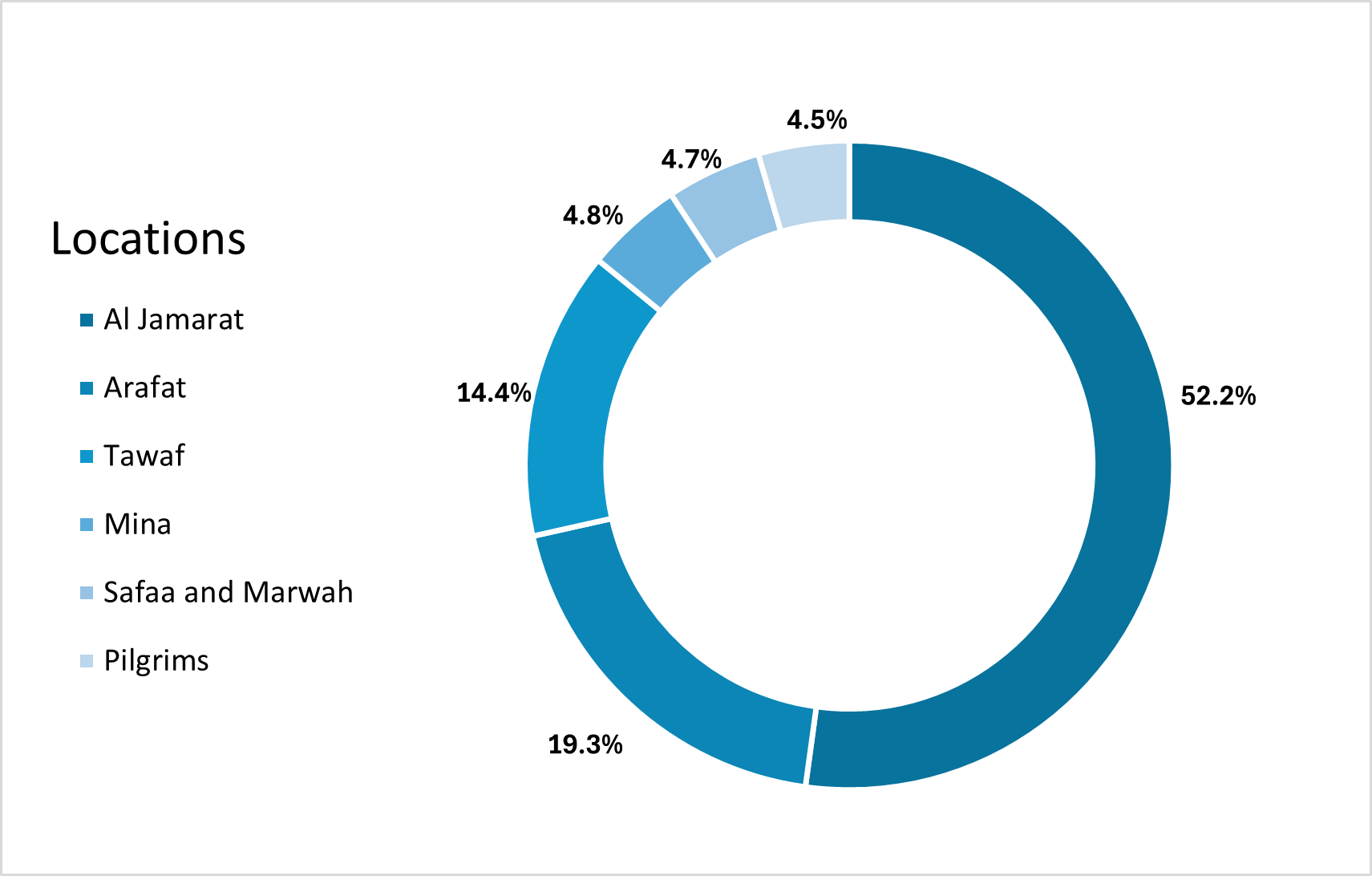}
\caption{Distribution of images per Location.}
\label{fig:locations}
\end{figure}

\FloatBarrier
Furthermore, our dataset contains frames from six main locations. Al-Jamarat provided the greatest number of frames at 52.2\%, followed by Arafat at 19.3\% of the dataset. The Tawaf area yielded 14.4\% of the total samples, while Mina, Safaa and Marwah, and the general route area accounted for 4.8\%, 4.7\%, and 4.5\%, respectively of the total frame samples. Fig.~\ref{fig:locations} shows the variations in the location captured in our dataset, which provides the model with an opportunity to be trained and generalized on various architectural configurations and crowd flow patterns.

\FloatBarrier

\begin{figure}[!htbp]
\centering
\includegraphics[width=\columnwidth, ,height=6cm]{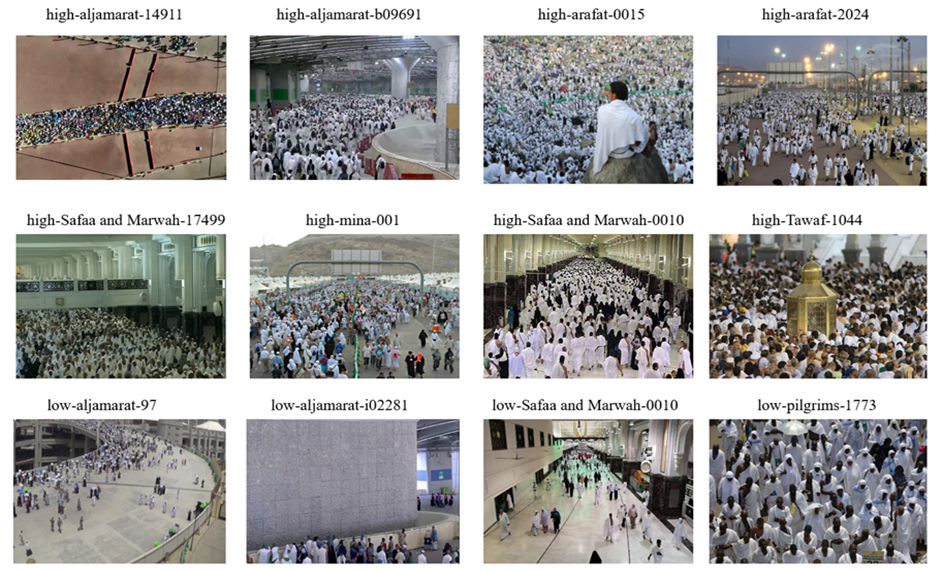}
\caption{Real data samples.}
\label{fig:samples}
\end{figure}

Representative examples from our dataset are illustrated in
Fig.~\ref{fig:samples}. The organization of the images is based on density and location. These samples underscore the quality and coverage of the dataset and the weak manual supervision labeling.

Dataset bias and representativeness considerations: The dataset covers multiple Hajj locations and density levels; however, sampling bias may remain because the frames were collected from publicly available YouTube videos rather than from systematically sampled operational or surveillance camera feeds. First, camera-angle selection bias may occur because public videos often emphasize visually clear, elevated, stable, or iconic views of ritual locations, while low-angle, obstructed, mobile-phone, and operational surveillance viewpoints may be underrepresented. Second, video-quality bias may arise because high-resolution, well-lit, and stable recordings are more likely to be selected and retained during preprocessing, whereas compressed, blurred, low-resolution, unstable, or extreme-illumination recordings may be less represented. Third, scene-composition bias may occur because some Hajj locations, rituals, crowd-flow patterns, and visually prominent viewpoints are more frequently recorded and uploaded than others. Consequently, the curated dataset and the generated CrowdH images may reflect the visual distribution of publicly available Hajj footage rather than the complete distribution of real deployment cameras. This limitation may affect the generalizability to operational monitoring settings with substantially different acquisition conditions.

Data leakage verification: All test datasets were strictly held out and were not used during training or model selection. For the in-domain HAJJv2 evaluation, the test split was sourced from YouTube video IDs that are not present in the training/validation video list, ensuring source-level disjointness. At the image level, we verified that there were zero exact overlaps between the training set (n = 993) and each evaluation set (HAJJv2 test; UCF-QNRF test; UCF\_CC\_50) using SHA-256 hashes \cite{b61} for exact-duplicate screening. We further checked for near-duplicates using perceptual hashing (pHash), confirming that the three test sets contained no duplicated or trivially re-encoded versions of training frames.
SHA-256 hash was computed on the stored image files (raw byte stream checksum), which detects only exact byte-level duplicates. Therefore, we additionally used perceptual hashing to detect near-duplicates under re-encoding, resizing, and compression. Using pHash (64-bit; imagehash.phash(hash\_size=8)), we confirmed zero exact and zero near-duplicate matches between the training set and the three test datasets under Hamming-distance thresholds of $\le 6$, $\le 10$, and $\le 16$(all yielded zero overlaps).

Dataset Partitioning: Our dataset consisted of 993 images.It was split into two sets: 792 images for training and 201 images for validation. We utilized this dataset partitioning to support model learning and ensure that the validation subset was well balanced. Furthermore, to evaluate how well the model performs when applied to different domains, we selected three open-source datasets that have been previously published as part of the testing phase. The HAJJv2 dataset \cite{b25}  was chosen as the domain-specific test data with 135 images extracted from the videos. In addition, two crowd counting benchmarks were used to evaluate the generalization performance of our model across different domains: UCF\_CC\_50 with 50 test images \cite{b14} and UCF-QNRF with 334 test images \cite{b4}. A summary of the dataset partitioning is provided in Table~\ref{tab:dataset_partition}.

\begin{table}[!htbp]

\caption{\textbf{Dataset composition and partitioning.}}
\label{tab:dataset_partition}
\centering

\footnotesize
\setlength{\tabcolsep}{2pt}
\renewcommand{\arraystretch}{1.25}
\setlength{\arrayrulewidth}{0.4pt}

\begin{tabularx}{\columnwidth}{|
p{0.22\columnwidth}|
p{0.30\columnwidth}|
>{\centering\arraybackslash}p{0.16\columnwidth}|
Y|}
\hline
Split & Source & No.\ of images & Purpose \\
\hline
Training &
\multirow{2}{0.30\columnwidth}{Our data (993 images)\\Split: 80\%--20\%} &
792 (train) &
Model training \\
\cline{1-1}\cline{3-4}
Validation &  &  201 (val) & Model validation \\
\hline
Test & HAJJv2 &  135 & In-domain evaluation \\
\hline
Test (cross-domain) & UCF-CC-50 & 50 & Generalization test set \\
\hline
Test (cross-domain) & UCF-QNRF & 334 & Generalization test set \\
\hline
\end{tabularx}
\end{table}

\subsection{PROPOSED MODEL DESIGN}
\begin{figure}[!htbp]
\centering
\includegraphics[width=\columnwidth]{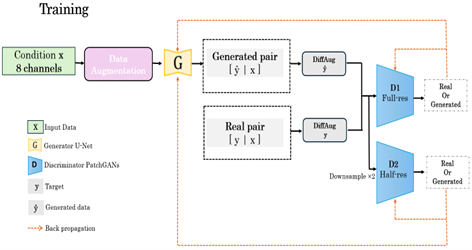}
\caption{Overview of the P2P-H training pipeline.}
\label{fig:pipeline}
\end{figure}

The proposed P2P-H architecture is illustrated in Fig.~\ref{fig:pipeline}. It integrates key components from P2P \cite{b7}, P2PHD \cite{b13}, and SGAN2-ADA \cite{b9} to improve structural preservation, semantic control, and training stability in crowd image synthesis. Unlike vanilla P2P, which uses a 3-channel RGB input and a single-scale PatchGAN discriminator, P2P-H introduces an eight-channel condition tensor for precise control and a multi-scale discriminator setup to enhance both local and global fidelity.

The pipeline begins with an eight-channel conditioning tensor   $x \in \mathbb{R}^{768 \times 768 \times 8}$, combining structural inputs (edges and grayscale) and semantic labels (crowd density and time-of-day one-hot encodings). This input undergoes optional spatial jitter and is processed by a deep U-Net generator G, which produces a synthetic image \(\hat{y}=G(x)\in[-1,1]^{768\times768\times3}\). Generated and real image-condition pairs \([\hat{y}\mid x]\) and \([y\mid x]\)
 were passed to two conditional PatchGAN discriminators \(\mathrm{D}_1\)  at full resolution (768×768) and \(\mathrm{D}_2\) at half resolution (384×384, via average pooling). Differentiable augmentations (DiffAug), including brightness, contrast, and spatial jitter, were applied only to image channels to regularize learning while preserving the conditioning semantics.

The generator was trained using a composite loss function comprising a hinge-based adversarial loss, feature-matching loss, $\ell_1$ pixel loss, and perceptual loss (VGG-19, \texttt{block3\_conv3}). To ensure smooth convergence and checkpoint stability, an exponential moving average (EMA) of the generator weights ($\beta$=0.999) was maintained throughout training. The full system was optimized using Two Time-Scale Update Rule (TTUR) Adam with staged loss scheduling and semantic dropout.

\subsubsection{Multi-Channel Conditioning Design}
The generator input layer is extended to accept the eight-channel conditioning tensor x described in the Dataset Labeling subsection, instead of a standard 3-channel RGB image. To prevent over-reliance on attribute channels, we applied label dropout with probability 0.12 to the semantic channels after epoch 40, while keeping structural channels intact.

\subsubsection{Generator Architecture}	
\begin{figure}[!htbp]
\centering
\includegraphics[width=\columnwidth]{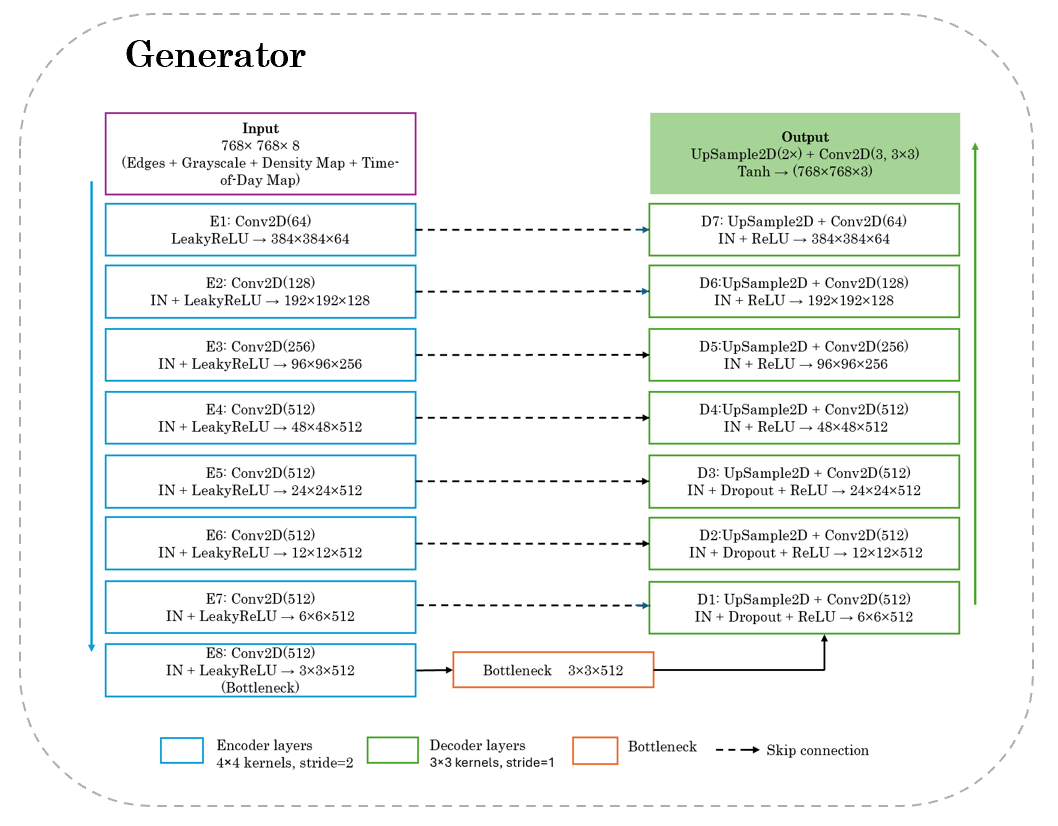}
\caption{Generator architecture (U-Net) of Pix2Pix-Hybrid. The network takes an 8-channel condition tensor (edges, grayscale, crowd-density labels, and time-of-day labels) and produces a 768×768 RGB image. Downsampling and upsampling blocks are connected by skipping connections to preserve spatial structure.}
\label{fig:Generator}
\end{figure}

The generator follows a deep U-Net structure comprising eight encoding and eight decoding layers, as shown in Fig.~\ref{fig:Generator}. The encoder uses 4×4 convolutions with stride 2, LeakyReLU activations ($\alpha$=0.2), and Instance Normalization in all blocks except the first. The decoder mirrors the encoder depth using nearest-neighbor upsampling with a scale factor of 2, followed by a 3×3 convolution, InstanceNorm, and ReLU activation. Dropout with a rate of 0.5 is applied to the first three decoder blocks. Skip connections concatenate the corresponding encoder features to each decoder layer to maintain spatial consistency. The final decoder layer employs an upsample followed by a 3×3 convolution with a tanh activation, yielding the final output \(\hat{y}\in[-1,1]^{768\times768\times3}\). This configuration avoids transposed convolutions, thus mitigating checkerboard artifacts in the synthesized images.

\subsubsection{Discriminator Architecture}	

\begin{figure}[!htbp]
\centering
\includegraphics[width=\columnwidth]{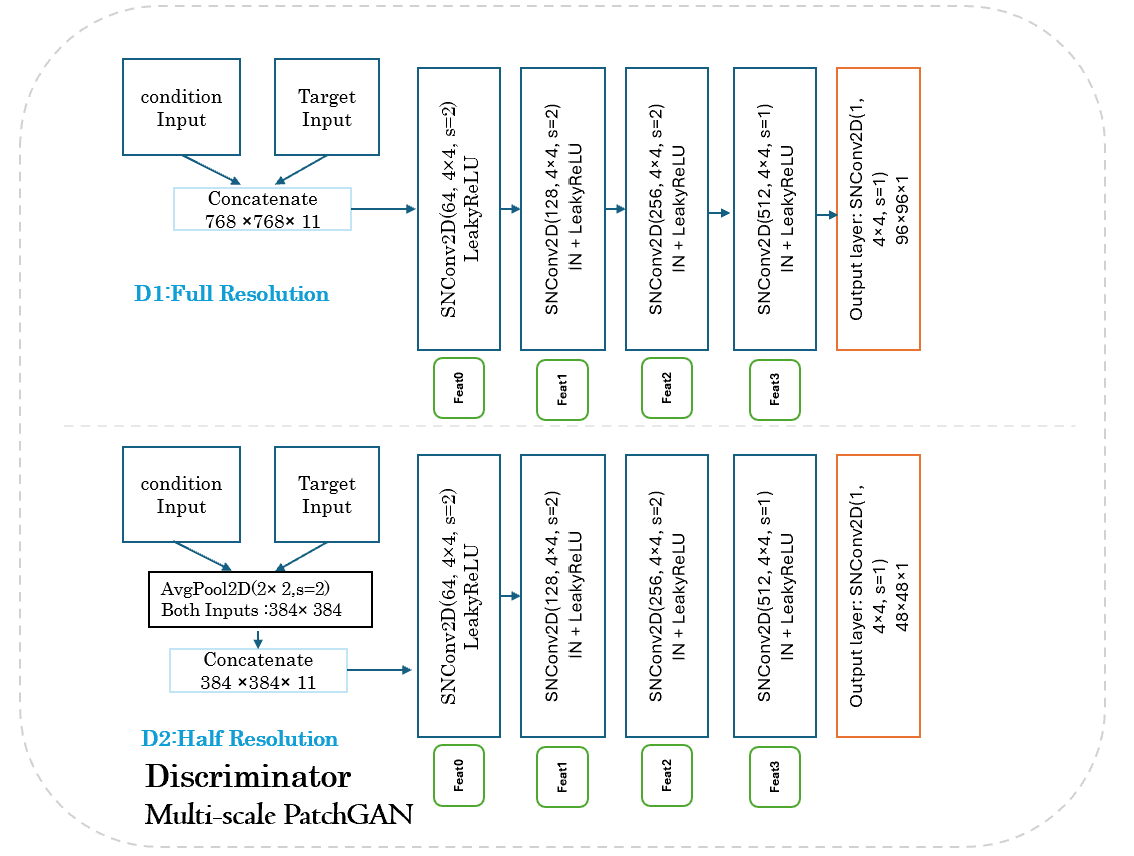}
\caption{Multi-scale PatchGAN discriminators. \(\mathrm{D}_1\) evaluates full-resolution condition–target pairs, while \(\mathrm{D}_2\) operates on inputs downsampled by 2×average pooling. Each discriminator backbone employs spectral normalization and outputs patchwise logit maps (96×96×1 and 48×48×1); intermediate feature maps are used in the feature-matching loss.}
\label{fig:Dis}
\end{figure}

Two PatchGAN discriminators were used to enforce both global and local consistency, as shown in Fig.~\ref{fig:Dis}: \(\mathrm{D}_1\) at full resolution (768×768) and \(\mathrm{D}_2\) at half resolution (384×384, obtained via average pooling). Each discriminator received concatenated eight-channel conditioning and three-channel RGB images, forming [x, y] or [x, ŷ] inputs. Both discriminators used five convolutional layers with 4×4 kernels, where the first three layers used stride 2 and the last two used stride 1. Spectral normalization was applied to all convolutional layers. LeakyReLU activation ($\alpha$= 0.2) and InstanceNorm (in all feature blocks except the first and final logit layers) were used throughout the experiment. The spatial output logits of the discriminators have sizes 96×96×1 for \(\mathrm{D}_1\) and 48×48×1 for \(\mathrm{D}_2\). Intermediate feature maps were extracted to compute the feature-matching loss. Differentiable augmentation included brightness adjustment, contrast scaling, and a ± 2-pixel spatial translation, applied only to image channels while preserving the conditioning integrity.An R1 gradient penalty ($\lambda$= 10.0) was applied to \(\mathrm{D}_1\)on real samples to stabilize training. This multi-scale architecture provides hierarchical feedback, improving both global coherence and fine-grained texture quality compared with the single-scale P2P baseline.

\subsection{MODEL TRAINING }
The proposed P2P-H model employed a multi-objective training strategy optimized via a composite loss function  as in~(\ref{eq:loss_total}):

\begin{equation}
\mathcal{L}_{\mathrm{total}}
= \lambda_{\mathrm{GAN}}\,\mathcal{L}_{\mathrm{adv}}
+ \lambda_{L1}\,\mathcal{L}_{L1}
+ \lambda_{\mathrm{perc}}\,\mathcal{L}_{\mathrm{perc}}
+ \lambda_{\mathrm{FM}}\,\mathcal{L}_{\mathrm{FM}} .
\label{eq:loss_total}
\tag{3}
\end{equation}

where:
\begin{itemize}
  \item \(\mathcal{L}_{\mathrm{adv}}\) is a hinge-based adversarial loss from two PatchGAN discriminators \((D_1, D_2)\), encouraging realism through dual-scale feedback.
  \item \(\mathcal{L}_{L1}\) is a pixel-level \(L1\) loss for low-frequency structural fidelity.
  \item \(\mathcal{L}_{\mathrm{perc}}\) is a perceptual loss computed on \texttt{block3\_conv3} of the VGG-19 feature maps of a pretrained VGG-19 network, enhancing semantic consistency.
  \item \(\mathcal{L}_{\mathrm{FM}}\) is a feature matching loss using intermediate features from both \(D_1\) and \(D_2\), promoting training stability and convergence.
\end{itemize}

Unlike vanilla P2P, which employs a single discriminator with a fixed loss weight, and P2PHD, which lacks dynamic loss adaptation, the proposed P2P-H framework implements a three-phase dynamic weighting schedule to ensure progressive learning. The three-phase dynamic weighting schedule, as shown in Table~\ref{tab:loss_schedule}. During epochs 1 to 10, training prioritized pixel-level reconstruction while minimizing adversarial influence. In epochs 11 to 40, the loss weights were adjusted to balance realism and fidelity. After epoch 40, the focus shifted to enhancing adversarial and feature-level losses to improve perceptual sharpness and the synthesis of fine details. The semantic conditioning channels were randomly zeroed with a dropout probability of p = 0.12 to encourage generalization. Each $\lambda$ (lambda) term represents the weight coefficient (or scaling factor) assigned to the loss component. These coefficients control the relative importance of each loss term during training.


\setlength{\tabcolsep}{4pt}
\renewcommand{\arraystretch}{1.25}
\begin{table}[!htbp]
\caption{Loss weight schedule across training epochs.}
\label{tab:loss_schedule}
\centering

\footnotesize
\setlength{\tabcolsep}{4pt}
\renewcommand{\arraystretch}{1.25}
\setlength{\arrayrulewidth}{0.4pt}

\begin{tabular}{|p{0.26\columnwidth}|c|c|c|c|c|}
\hline
Epoch Range & $\lambda_{L1}$ & $\lambda_{\mathrm{perc}}$ & $\lambda_{\mathrm{GAN}}$ & $\lambda_{\mathrm{FM}}$ & Dropout ($p$) \\
\hline
1--10  & 60.0 & 1.0 & 1.0 & 5.0 & 0.0  \\
\hline
11--40 & 40.0 & 0.5 & 1.0 & 5.0 & 0.0  \\
\hline
41+    & 40.0 & 0.5 & 1.1 & 6.0 & 0.12 \\
\hline
\end{tabular}
\end{table}

Model optimization followed TTUR with separate learning rates for the generator (G) and the discriminators (D). The details are summarized in Table~\ref{tab:opt_hparams}, where LR = learning rate, Optim. = optimizer, Grad. Clip = gradient clipping norm, EMA = Exponential Moving Average factor.

The model used an exponential moving average (EMA) of the generator weights to evaluate and select checkpoints. This training method, which includes multi-scale supervision, adaptive loss scheduling, and semantic regularization, helps generate high-quality, realistic crowd images.

\begin{table}[!htbp]
\caption{Optimizer and training hyperparameters.}
\label{tab:opt_hparams}
\centering

\footnotesize
\setlength{\tabcolsep}{2pt}
\renewcommand{\arraystretch}{1.15}
\setlength{\arrayrulewidth}{0.4pt}

\begin{tabular*}{\columnwidth}{@{\extracolsep{\fill}}|c|c|c|c|c|c|c|c|}
\hline
Comp. & LR & Optim. & $\beta_1$ & $\beta_2$ &
\shortstack[c]{Grad.\\Clip.} &
\shortstack[c]{EMA\\($\beta$)} &
\shortstack[c]{LR Decay\\($\gamma$)} \\
\hline
$G$ & $1\times10^{-4}$ & Adam & 0.5 & 0.999 & 1.0 & 0.999 & 0.95 \\
\hline
\shortstack[c]{$D_1,$\\$D_2$} & $2\times10^{-4}$ & Adam & 0.5 & 0.999 & 1.0 & -- & 0.95 \\
\hline
\end{tabular*}
\end{table}

\FloatBarrier

\subsection{MODEL TESTING AND EVALUATION}

The proposed Pix2Pix-Hybrid model was evaluated to assess both in-domain and cross-domain  performance, as summarized in Table 2. In-domain evaluation was conducted on the HAJJv2 dataset \cite{b25}  using unseen held-out samples under domain-consistent conditions. Cross-domain generalization was assessed on UCF-QNRF \cite{b4} and UCF\_CC\_50 \cite{b14}, which exhibit different scene characteristics. Quantitative evaluation is organized into (i) fidelity/perceptual consistency metrics and (ii) distribution-level similarity metrics, complemented by qualitative consistency analysis.

\subsubsection{Quantitative Evaluation}
There is no single metric that captures all facets of generative model performance \cite{b46}. Therefore, several measures were used to quantify fidelity, perceptual similarity, and distributional consistency. 

Fr\'echet Inception Distance (FID): FID computes the Fr\'echet (Wasserstein-2) distance
between multivariate Gaussians fitted to Inception feature embeddings of real and generated
images \cite{b47}, \cite{b48}. FID is defined in~(\ref{eq:fid}),
where \((\boldsymbol{\mu}_r,\mathbf{\Sigma}_r)\) and \((\boldsymbol{\mu}_g,\mathbf{\Sigma}_g)\) denote the
empirical mean and covariance of the real and generated features, respectively.

\begin{equation}
\mathrm{FID}
= \left\lVert \boldsymbol{\mu}_r - \boldsymbol{\mu}_g \right\rVert_2^2
+ \mathrm{Tr}\!\left(
\mathbf{\Sigma}_r + \mathbf{\Sigma}_g
- 2\left(\mathbf{\Sigma}_r \mathbf{\Sigma}_g\right)^{\tfrac{1}{2}}
\right).
\label{eq:fid}
\tag{4}
\end{equation}

A lower FID indicates closer alignment between real and generated feature distributions. FID is known to be sensitive to the number of samples used for estimation, motivating the
inclusion of KID \cite{b49}.

Kernel Inception Distance (KID): KID measures distributional similarity using an unbiased estimator of the squared Maximum Mean Discrepancy (MMD) in Inception feature space with a polynomial kernel \cite{b50}.
Given real features $X_r$ and generated features $X_g$, KID is computed as shown in (5):

\begin{equation}
\text{KID} = \text{MMD}^2(X_r, X_g)
\label{eq:kid}
\tag{5}
\end{equation}

where the squared Maximum Mean Discrepancy (MMD) is computed as in (6):

\begin{align}
\text{MMD}^2(X_r, X_g) &= \mathbb{E}_{x,x' \sim X_r}[k(x,x')] \nonumber \\
&\quad - 2\mathbb{E}_{x \sim X_r, y \sim X_g}[k(x,y)] \nonumber \\
&\quad + \mathbb{E}_{y,y' \sim X_g}[k(y,y')]
\label{eq:mmd}
\tag{6}
\end{align}
where $x$ and $x'$ are samples from the real distribution, $y$ and $y'$ are samples from the generated distribution.Here,the polynomial kernel  k(a,b) is the kernel function evaluating the similarity between samples  a and b as in (7)
\begin{equation}
k(a, b) = \left( \frac{1}{d} a^T b + 1 \right)^3
\label{eq:kernel}
\tag{7}
\end{equation}

In our implementation, we use $d$ as the dimensionality of the feature vectors. Lower KID indicates better distributional agreement \cite{b50}. In this work, KID is reported as the mean $\pm$ standard deviation over multiple random subsets.

Structural Similarity Index Measure (SSIM): SSIM evaluates local structural similarity between a generated
image $y$ and its paired reference $x$ using luminance ($l$), contrast ($c$), and structure ($s$) terms \cite{b51}. The individual comparison functions are as in (8), (9), and (10):  

\begin{equation}
l(x, y) = \frac{2\mu_x \mu_y + c_1}{\mu_x^2 + \mu_y^2 + c_1}
\tag{8}
\end{equation}

\begin{equation}
c(x, y) = \frac{2\sigma_x \sigma_y + c_2}{\sigma_x^2 + \sigma_y^2 + c_2}
\tag{9}
\end{equation}

\begin{equation}
s(x, y) = \frac{\sigma_{xy} + c_3}{\sigma_x \sigma_y + c_3}
\tag{10}
\end{equation}
where $\mu_x$ and $\mu_y$ denote local means, $\sigma_x^2$ and $\sigma_y^2$ denote local variances,
$\sigma_{xy}$ is the local covariance, and $C_1,C_2$ are stabilizing constants \cite{b51}.
The SSIM for each block is then a weighted combination of those comparative measures as formed in (11):
\begin{equation}
\text{SSIM}(x, y) = l(x, y)^\alpha \cdot c(x, y)^\beta \cdot s(x, y)^\gamma
\tag{11}
\end{equation}

Peak Signal-to-Noise Ratio (PSNR): Measures the PSNR between two monochrome images $I$ and $K$  to assess the quality of a generated image compared to its corresponding real image. It is derived from the mean squared error (MSE) \cite{b50}, \cite{b51}. Higher PSNR (measured in dB) indicates better reconstruction quality, with the generated image being closer to the reference image. It is computed as in (12), (13), and (14).

\begin{equation}
\mathrm{PSNR}(I,K)
= 10\log_{10}\!\left(\frac{MAX_I^2}{\mathrm{MSE}}\right).
\label{eq:psnr}
\tag{12}
\end{equation}

\begin{equation}
\begin{aligned}
&= 20\log_{10}(\mathrm{MAX}_I) - 10\log_{10}(\mathrm{MSE}(I,K)).
\end{aligned}
\label{eq:psnr}
\tag{13}
\end{equation}
Where 
\begin{equation}
\mathrm{MSE}(I,K)
= \frac{1}{mn}\sum_{i=0}^{m-1}\sum_{j=0}^{n-1}\big(I(i,j)- K(i,j)\big)^2 .
\label{eq:mse_gray}
\tag{14}
\end{equation}
where $\mathrm{MAX}$ is the maximum possible pixel value of the image and N is the number of pixels.

Learned Perceptual Image Patch Similarity (LPIPS): LPIPS measures perceptual similarity using deep features from a pretrained network, such as VGG and correlates well with human judgments \cite{b52}. A lower LPIPS indicates perceptually closer reconstructions.

Metric interpretation under self-paired conditional training. In our self-paired setup, each output is reconstructed from an edge-derived representation of the same target image. Therefore, SSIM and PSNR primarily measure conditional reconstruction fidelity (edge-to-image faithfulness).High SSIM and PSNR values are expected because the objective is to reproduce the paired target rather than generate unconstrained, diverse samples.We complement SSIM and PSNR with LPIPS, which captures paired perceptual similarity using deep feature distances. In contrast, FID and KID are distribution-level measures computed on held-out data splits. These metrics quantify whether outputs produced from unseen conditioning inputs match the real-image feature statistics, providing supporting evidence of conditional realism and generalization beyond pixel-wise reconstruction.
\subsubsection{Evaluation Protocol}
All metrics were computed on held-out VAL ($n=201$) and TEST ($n=135$) at
$768\times 768$ output resolution. For each real target image $y$, we constructed
the conditioning tensor $x=g(y)$ (edges + grayscale + one-hot density/time) and
generated $\hat{y}=G(x)$. Metrics are then computed between the real $y$ and the
generated $\hat{y}$. We fixed the random evaluation seed to \texttt{1337} for
NumPy/TensorFlow, and we used \texttt{SPLIT\_SEED=2025} for the pinned split file
\texttt{val.txt}. For KID, subset sampling used \texttt{seed=1337}.

Fidelity metrics (SSIM/PSNR/LPIPS): Generated and real RGB images were represented in $[-1,1]$.
For SSIM and PSNR, both images were mapped to $[0,1]$ via (x+1) / 2 with clipping and evaluated using TensorFlow and then averaged over all images in the split. LPIPS was computed using the LPIPS (VGG) implementation on tensors in the $[-1,1]$ range (\texttt{NCHW} format) and reported as the mean over the split.

Distribution metrics (FID/KID): We computed FID and KID using InceptionV3 network architecture (\texttt{include\_top
=False}, \texttt{pooling=avg}, ImageNet weights) implemented in TensorFlow/Keras v2.19.0, extracting 2048-D pooled features for both real and generated images. Images were converted from [-1,1] to [0,255] as float32, resized to 299×299 using bilinear interpolation with antialiasing, then preprocessed using the InceptionV3 preprocessing function prior to feature extraction. KID was calculated  as an unbiased estimate of MMD² in Inception feature space using a polynomial kernel. KID was computed 50 times, each time by randomly drawing K real and k generated feature vectors (without replacement), producing 50 KID values. Then we reported their mean ± standard deviation as the final KID for that split. The subset size as in (15):
\begin{equation}
k = \min\!\left(1000,\; \min\!\left(n_{\mathrm{real}},\, n_{\mathrm{gen}}\right) - 1\right)
\label{eq:k_subset}
\tag{15}
\end{equation}

which yielded (k=200) on VAL (n=201) and k=134 on TEST (n=135) in our evaluation. We fixed the random seed to 1337 for the reproducibility of the subset sampling.
\subsubsection{Cross-Domain Generalization}
We conducted a cross-domain evaluation to test our model generalization using the UCF\_CC\_50 \cite{b14} and UCF-QNRF \cite{b4} datasets. These two open-source datasets were selected for several reasons. The UCF\_CC\_50 dataset provides extreme density variations, whereas the UCF-QNRF dataset offers large-scale diversity. These two corpora together test the model robustness across different crowd conditions and contain general crowd scenes that are distinct from Hajj-specific images. This evaluation tests whether our model learns transferable crowd features. To determine whether our model learned transferable crowd representations, we compared four performance metrics: FID, KID, SSIM, and LPIPS. The testing methodology allowed us to assess how well our model generalized to unseen crowd distributions and environmental conditions.

\subsection{SYNTHETIC DATASET GENERATION VALIDATION}
After completing model training and validation, we deployed the best-performing checkpoint and produced 10,000 synthetic Hajj crowd images at a resolution of 768×768 pixels. The images produced were then evaluated to assess distributional similarity, appearance-level variation, and distinctness from the training samples.

In addition to the numerical analysis, we used qualitative assessments to examine the visual plausibility and diversity of the generated images. We also trained a binary convolutional neural network (CNN) classifier to distinguish real from synthetic samples \cite{b53}. This classifier was used as an auxiliary real–synthetic separability check under a specific classifier protocol.
Five training/testing splits (40\%, 50\%, 60\%, 70\%, 80\%)  were examined. Four metrics were measured for each configuration. A true positive (TP) occurs when a real image is correctly classified as real. False negatives (FN) occur when real images are misclassified as synthetic. True negative (TN) occurs when synthetic images are correctly classified as synthetic. False positive (FP) counts synthetic images incorrectly classified as real. From these values, we calculated the following metrics using Equations (16), (17), (18), and (19).

\setcounter{equation}{15} 

\begin{IEEEeqnarray}{lCl}
\mathrm{Accuracy}  & = & \frac{TP+TN}{TP+FP+TN+FN} \label{eq:acc} \\
\mathrm{Precision} & = & \frac{TP}{TP+FP} \label{eq:prec} \\
\mathrm{Recall}    & = & \frac{TP}{TP+FN} \label{eq:rec} \\
\mathrm{F1\text{-}Score} & = & \frac{2\,\mathrm{Precision}\,\mathrm{Recall}}
{\mathrm{Precision}+\mathrm{Recall}} \label{eq:f1}
\end{IEEEeqnarray}

To further examine feature-space organization, t-distributed stochastic neighbor embedding (t-SNE) \cite{b56} was used to project high-dimensional Inception features into a two-dimensional space. The t-SNE visualizations were used as qualitative supporting evidence for real–synthetic feature-space organization and attribute-related trends.

To assess whether the P2P-H generator reproduces training images, we conducted a nearest-neighbor (NN) memorization check in feature space using a retrieval-then-distance protocol \cite{b50}. Specifically, we embedded all training images and each generated sample using CLIP ViT-B/32 image features \cite{b62}. We retrieved the top-5 nearest training neighbors using cosine similarity in the CLIP embedding space.For each generated image, we then computed the LPIPS perceptual distance (VGG backbone) between each generated sample and each of its retrieved neighbors \cite{b52}. We also reported NN1 as the LPIPS distance to the closest retrieved neighbor (the highest-similarity CLIP match). LPIPS provides a perceptually aligned similarity measure that is more informative than pixel-level metrics for detecting near-duplicate reproduction. In addition, we computed a Train→Train baseline by measuring the LPIPS distance from each training image to its nearest neighbor within the training set. This baseline quantified the expected level of natural near-duplication among real frames (e.g., consecutive video frames) and provides a conservative reference distribution for comparison.Finally, we reported (i) the NN1 LPIPS distribution (Generated→Train compared to Train→Train), (ii) ECDF(Empirical Cumulative Distribution Function) and histogram overlays, and (iii) representative qualitative grids for worst-case samples(lowest NN1-LPIPS) and grids for random samples as[Generated | NN1–NN5] annotated with similarity scores for visual inspection. The same NN+LPIPS protocol was applied using generated outputs conditioned on three test datasets: HAJJv2, UCF\_CC\_50, and UCF-QNRF.

\subsection{Crowd-Counting Validation on CrowdH-Mix-469}

\subsubsection{Synthetic Data Selection and Statistical Analysis of CrowdH-Mix-469}
To construct CrowdH-Mix-469, we first applied an automated cleaning process to the 10,000 synthetic images in CrowdH. Duplicate removal was performed automatically, as this step is straightforward to justify and helps eliminate exact and near-duplicate synthetic samples. In contrast, quality-based filtering was applied more cautiously and was supported by manual review to avoid discarding visually difficult but still potentially useful crowd scenes. Such cases included dark, bright, and mildly blurred images, which may still reflect realistic acquisition conditions. After this cleaning stage, 7,000 unique usable synthetic samples were retained. From this cleaned image pool, 85 synthetic images were selected in a balanced manner across crowd density, scene/location, viewpoint, and illumination conditions. After final manual verification, the selected images were annotated with head points using the VIA tool and incorporated into CrowdH-Mix-469 \cite{b65}.

CrowdH-Mix-469 used in this study comprises 469 images, including 384 real Hajj images and 85 high-resolution synthetic images generated by the proposed P2P-H framework, as summarized in Table~\ref{tab:crowdh_mix_469_compare}. Among the real images, 307 were used for training and 77 were reserved for testing, while the synthetic images were used only for training enhancement. This protocol ensures that all reported crowd-counting results were evaluated exclusively on held-out real Hajj images, thereby preserving evaluation integrity and avoiding overlap between the generative and downstream counting stages. Compared with representative real and synthetic crowd-counting datasets, CrowdH-Mix-469 is smaller in scale but more closely aligned with the target Hajj domain, which remains underrepresented in existing public benchmarks. It also maintains relatively high image resolution and covers a broad range of crowd densities, making it suitable for studying dense and structurally complex pilgrimage scenes.


\begin{table}[!htbp]
\caption{Statistical comparison of CrowdH-Mix-469 with representative real and synthetic crowd-counting datasets.}
\label{tab:crowdh_mix_469_compare}
\centering

\scriptsize
\setlength{\tabcolsep}{1pt}
\renewcommand{\arraystretch}{1.15}
\setlength{\arrayrulewidth}{0.4pt}

\newcolumntype{C}[1]{>{\centering\arraybackslash}p{#1}}
\newcolumntype{L}[1]{>{\raggedright\arraybackslash}p{#1}}

\begin{tabular}{|L{0.18\columnwidth}|C{0.18\columnwidth}|C{0.11\columnwidth}|C{0.14\columnwidth}|C{0.07\columnwidth}|C{0.07\columnwidth}|C{0.07\columnwidth}|C{0.08\columnwidth}|}
\hline
\textbf{Data set} &
\textbf{\shortstack{Avg. img.\\size}} &
\textbf{\# Images} &
\multicolumn{4}{c|}{\textbf{Pedestrian count}} &
\textbf{Type} \\
\cline{4-7}
 & & & \textbf{Total} & \textbf{Min} & \textbf{Avg} & \textbf{Max} & \\
\hline
GCC \cite{b28} & 1080$\times$1920 & 15,212 & 7,625,843 & 0 & 501 & 3995 & Synth \\
\hline
CVCS \cite{b30} & 1080$\times$1920 & 280,000 & -- & 90 & -- & 180 & Synth \\
\hline
CrowdX \cite{b29} & 1024$\times$768 & 24,000 & -- & 1 & 500 & 1000 & Synth \\
\hline
SHT A \cite{b3} & 589$\times$868 & 482 & 241,677 & 33 & 501 & 3139 & Real \\
\hline
SHT B \cite{b3} & 768$\times$1024 & 716 & 88,488 & 9 & 123 & 578 & Real \\
\hline
WorldExpo'10\cite{b17} & 576$\times$720 & 3980 & 199,923 & 1 & 50 & 253 & Real \\
\hline
UCF-QNRF \cite{b4} & 2013$\times$2902 & 1525 & 1,251,642 & 49 & 815 & 12,865 & Real \\
\hline
NWPU \cite{b19} & 2191$\times$3209 & 5109 & 2,133,375 & 0 & 418 & 20,033 & Real \\
\hline
CrowdH-Mix-469 (our) & 1025$\times$1482 & 469 & 233,828 & 10 & 499 & 5665 & Mixed \\
\hline
\end{tabular}
\end{table}

\subsubsection{Crowd-Counting Validation Protocol}
To assess the practical utility of the generated images for downstream crowd analysis, five representative crowd-counting models were selected: MCNN, CSRNet, DM-Count, P2PNet, and APGCC. These models span several architectural families: MCNN \cite{b3} as a classical multi-column CNN baseline, CSRNet \cite{b2} as a dilated-convolution density-regression model, DM-Count \cite{b66} as an optimal-transport-based density-regression method, P2PNet \cite{b67} as a point-based localization and counting framework, and APGCC \cite{b68} as a recent point-based crowd-counting and localization framework based on auxiliary point guidance.

Each model was evaluated under two training settings: (i) training with real Hajj data only (n=307) and (ii) training with the same real Hajj data augmented with 85 selected synthetic images generated by the proposed P2P-H framework. In all cases, evaluation was performed exclusively on the held-out real Hajj test set (n = 77), so that the downstream assessment remained independent of the synthetic generation stage. Within each model family, the real-only and real-plus-synthetic experiments were conducted under the same training configuration to ensure a fair comparison and to attribute performance differences to the training data rather than optimization changes. Performance was measured using Mean Absolute Error (MAE), Mean Squared
Error (MSE), and Root Mean Squared Error (RMSE) \cite{b2}. Each model was initialized from publicly available pretrained crowd-counting weights and then fine-tuned on the target Hajj training split under the two settings described above.

\section{Experimental Results and Discussion}

To assess the effectiveness of our approach, we compared it with two baseline models on the Hajj crowd dataset: (i) the standard P2P conditional GAN \cite{b7} and (ii) SGAN2-ADA as an unconditional model \cite{b9}. Both baselines were trained using their default hyperparameters. For P2P, because the data lacked paired labels, we paired the RGB images with their corresponding derived edge maps by applying a 5 × 5 Gaussian blur with sigma=1.0 and using Canny edge detection at thresholds of 64 and 128. All input pairs were normalized to the range [-1,1] to stabilize the training.

P2P serves as the principal task-matched baseline, because it operates in the same conditional image-to-image translation setting as the proposed method. By contrast, SGAN2-ADA was included only as an exploratory unconditional reference. In practice, the unconditional SGAN2-ADA formulation was unable to preserve the scene structure and crowd layout required in this problem setting, whereas the standard P2P baseline still suffered from blurred textures and checkerboard artifacts due to its simpler conditioning design and transposed-convolution-based decoder. These observations motivated the development of the proposed P2P-H framework, which was designed specifically for structure-guided Hajj crowd synthesis. These qualitative shortcomings were further quantified by fidelity, generalization, and qualitative analyses reported below.

\subsection{PROPOSED P2P-H MODEL EVALUATION}

\subsubsection{Quantitative Evaluation}

\begin{table*}[!b]
\caption{Quantitative comparison of P2P, SGAN2-ADA, and the proposed P2P-H across FID, KID, SSIM, PSNR, and LPIP On the validation and HAJJv2 test split.}
\label{tab:quantitative_comparison}
\centering

\footnotesize
\setlength{\tabcolsep}{3pt}
\renewcommand{\arraystretch}{1.15}
\setlength{\arrayrulewidth}{0.4pt}

\begin{tabular}{|l|l|c|c|c|c|c|c|}
\hline
Model & Split & SSIM $\uparrow$ & PSNR (dB) $\uparrow$ & LPIPS $\downarrow$ & FID $\downarrow$ & KID $\downarrow$ & Best epoch \\
\hline
\multirow{2}{*}{Standard P2P (HAJJv2)} & Val ($n{=}201$)  & 0.0099 & 5.18 & 0.7499 & 440.54 & $0.450 \pm 0.025$ & \multirow{2}{*}{89} \\
\cline{2-7}
 & Test ($n{=}135$) & 0.0072 & 5.09 & 0.7315 & 513.61 & $0.530 \pm 0.035$ & \\
\hline
\multirow{2}{*}{StyleGAN2-ADA (HAJJv2)} & Val ($n{=}201$)  & 0.1400 & 10.40 & 0.3420 & 133.73 & $0.115 \pm 0.008$ & \multirow{2}{*}{500} \\
\cline{2-7}
 & Test ($n{=}135$) & 0.1247 & 9.99 & 0.3120 & 268.01 & $0.252 \pm 0.018$ & \\
\hline
\multirow{2}{*}{Proposed P2P-H (HAJJv2)} & Val ($n{=}201$)  & 0.9341 & 26.95 & 0.1851 & 59.26  & $0.0008 \pm 0.000097$ & \multirow{2}{*}{116} \\
\cline{2-7}
 & Test ($n{=}135$) & 0.9610 & 30.37 & 0.1332 & 48.01  & $0.0023 \pm 0.0002$ & \\
\hline
\end{tabular}
\end{table*}
The comparison of quantitative evaluation results is shown in Table~\ref{tab:quantitative_comparison}. To ensure fair evaluation, all models were completely trained, and the best checkpoint was selected based on the lowest FID on the validation set. The proposed model showed better performance than both baselines across all fidelity metrics. On the validation set (n=201), it achieved SSIM = 0.9341, PSNR = 26.95 dB, LPIPS = 0.1851, FID = 59.26, and KID = -0.0008 ± 0.000097. These values indicate that P2P-H achieved strong similarity to real images and high reconstruction fidelity, supporting successful edge-to-image reconstruction under the proposed conditional framework. In addition, the results of the proposed model improved when tested on the HAJJv2 dataset, achieving SSIM = 0.9610, PSNR = 30.37 dB, LPIPS = 0.1332, FID = 48.005, and KID = 0.0023 ± 0.0002. These results indicate strong structural similarity (high SSIM and PSNR), good perceptual quality (low LPIPS), and a close match between the real and synthetic feature distributions (low FID and KID). These results reflect strong in-domain performance on held-out HAJJv2 inputs. This indicates its ability to reconstruct realistic crowd images that remain structurally consistent with the conditioning input while preserving crowd appearance.

On the other hand, the standard P2P model with only edge-based conditioning achieved poor results on HAJJv2. It reached SSIM = 0.0072, FID = 513.61, and PSNR = 5.09 dB.These results indicate the inability of P2P to reconstruct complex crowd textures from limited structural information. The proposed model addresses this issue by using a multi-channel conditioning input that combines structural guidance and contextual attributes to generate semantically and visually consistent outputs. SGAN2-ADA improved over the standard P2P model, as reflected by SSIM = 0.1400, LPIPS = 0.3420, FID = 133.73, and KID = 0.115 ± 0.008 on the validation set. This improvement is due to the benefits of adaptive discriminator augmentation in small-data regimes. However, its unconditional formulation prevents it from maintaining the spatial layout specific to each scene. Consequently, FID and structural fidelity were worse than those of the proposed model.

\subsubsection{Qualitative Assessment}
\begin{figure*}[!p]
\centering

\includegraphics[width=\textwidth,height=0.30\textheight]{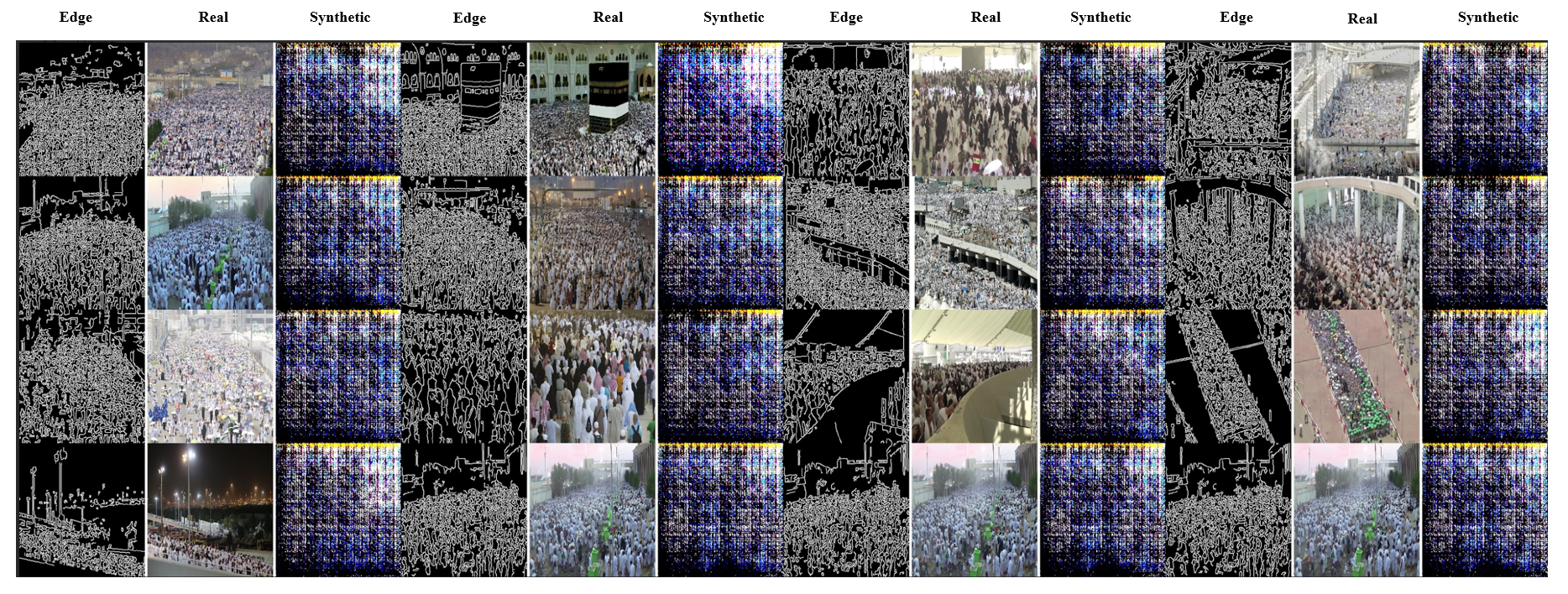}
\captionof{figure}{Samples generated from the standard P2P model.}
\label{fig:P2P_samples}

\vspace{3pt}

\includegraphics[width=\textwidth,height=0.31\textheight]{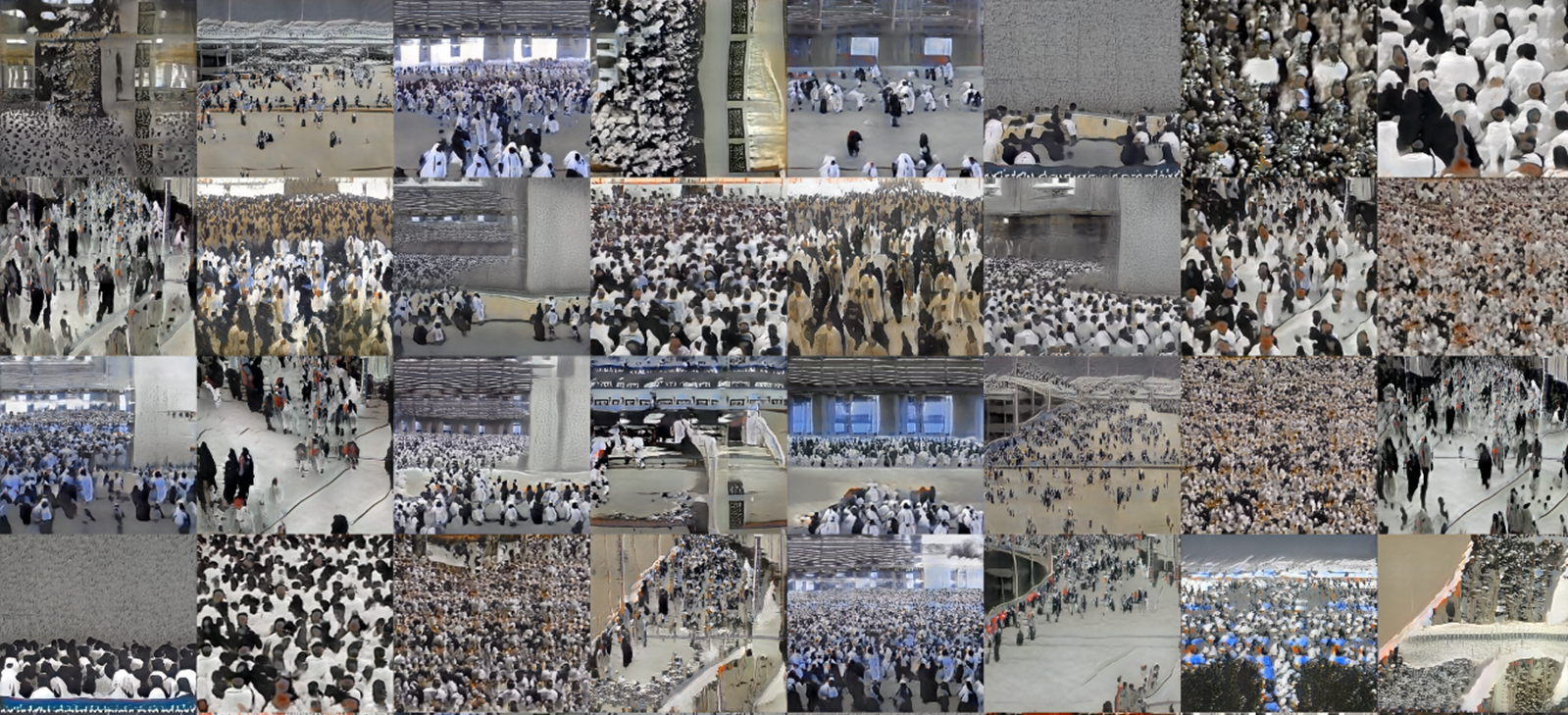}
\captionof{figure}{Samples generated from StyleGAN2-ADA model.}
\label{fig:SGAN-ADA_samples}

\vspace{3pt}

\begin{subfigure}{\textwidth}
  \centering
  \includegraphics[width=\textwidth,height=0.30\textheight]{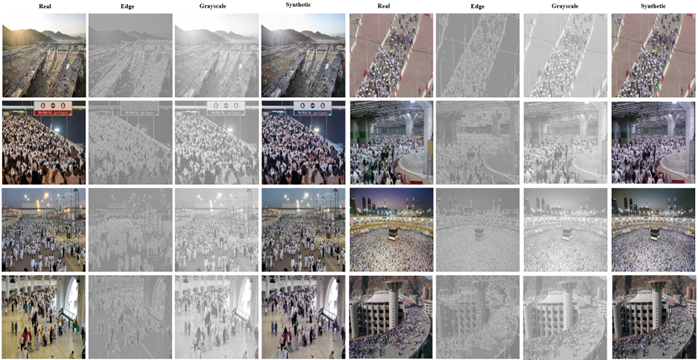}
  \caption{Samples generated from our proposed  P2P-H model.}
  \label{fig:P2P-H_samples}
\end{subfigure}

\end{figure*}

In addition to quantitative measurements, a comprehensive qualitative assessment was performed to visually examine the realism, structural fidelity, and attribute responsiveness of the generated images. Fig.~\ref{fig:P2P_samples} illustrates the samples produced using the standard P2P model. This baseline was trained using edge-to-image translation. The model captures the overall scene layout. However, several quality issues appear in the model outputs. Checkerboard artifacts are visible throughout the outputs. Furthermore, the texture diversity was very low compared with real-world images. Additionally, illumination was inconsistent across different samples.

Fig.~\ref{fig:SGAN-ADA_samples} presents samples generated by SGAN2-ADA trained on the Hajj dataset with adaptive discriminator augmentation. Some individual images appear realistic. However, several shortcomings were apparent. The diversity of crowd configurations was limited, with indications of mode collapse and repeated patterns in dense areas. In general, the complex architectural structures and geometric details surrounding important ritual locations were either poorly defined or highly distorted. In high-density areas, pilgrims often appeared merged or heavily deformed, and figure boundaries were unclear. These issues are consistent with the unconditional nature of SGAN2-ADA, which provides no explicit control over scene attributes such as crowd density, time of day, or camera viewpoint, making it difficult to enforce spatial and semantic consistency.

Fig.~\ref{fig:P2P-H_samples} displays the samples generated by the proposed P2P-H model. Various Hajj crowd scenes were synthesized under different structural and contextual conditions. The model

\FloatBarrier

 \FloatBarrier

\FloatBarrier

performed well across different environments. Both outdoor and indoor scenes were reconstructed with realistic appearance and appropriate detail. Multiple visual quality indicators were positive. The structural details remained consistent throughout the generated images. The geometry of the building and the architectural elements look accurate. Crowd formations seem natural rather than artificial, and people are positioned realistically. The dense regions did not collapse into blurred masses. Lighting responded appropriately to the input conditions. The color and illumination accurately reflect the specified times of the day. These observations support the effectiveness of conditional generation.

Together, Figs. 8-10 visually support the quantitative results reported in Table~\ref{tab:quantitative_comparison} by comparing the baseline outputs with the visually plausible, structure-preserving outputs produced by P2P-H under fixed structural conditioning. These results are further discussed in Section IV-B.

\begin{table*}[!b]
\caption{Progressive component-integration experiments for the proposed P2P-H framework. Unless otherwise stated, all experiments use the same self-paired train/validation split, the same U-Net generator backbone, 768$\times$768 output resolution, and the same $\ell_1$ reconstruction term.}
\label{tab:progressive_integration}
\centering
\footnotesize
\setlength{\tabcolsep}{3pt}
\renewcommand{\arraystretch}{1.15}
\setlength{\arrayrulewidth}{0.4pt}

\begin{tabular}{|c|p{2.0cm}|p{2.0cm}|p{3.0cm}|p{3.2cm}|p{4.6cm}|}
\hline
\textbf{Exp.} &
\textbf{Experiment name} &
\textbf{Conditioning input} &
\textbf{Discriminator setup} &
\textbf{Newly introduced components at this stage} &
\textbf{Purpose of the stage} \\
\hline
\textbf{E0} & Initial Weak labels & Edge + Density + time-of-day & Single PatchGAN &
Weak semantic labels under single-discriminator training &
Provides the initial hybrid baseline before grayscale and advanced stabilization are introduced \\
\hline
\textbf{E1} & Improved weak-label baseline & Density + time-of-day & Single PatchGAN &
Hinge adversarial loss + perceptual loss &
Examines whether basic loss-level refinements improve the initial weak-label baseline without changing the conditioning design \\
\hline
\textbf{E2} & Hybrid-conditioning baseline & Edge + Density + time-of-day & Single PatchGAN &
Full multi-channel conditioning introduced &
Isolates the contribution of the proposed hybrid conditioning design under a single-discriminator setting \\
\hline
\textbf{E3} & Stabilized hybrid single-D model & Edge + grayscale + density + time & Single PatchGAN &
Perceptual loss, feature matching, spectral normalization, R1, EMA &
Assesses whether stronger stabilization improves the hybrid-conditioning model before multi-scale discrimination is added \\
\hline
\textbf{E4} & Multi-scale hybrid model & Edge + grayscale + density + time &
Two PatchGAN discriminators (full + half resolution) &
Multi-scale discrimination + image-only DiffAug &
Measures the contribution of hierarchical adversarial feedback and image-only augmentation once hybrid conditioning is fixed \\
\hline
\textbf{E5} & Final P2P-H model & Edge + grayscale + density + time &
Two PatchGAN discriminators (full + half resolution) &
Dynamic loss schedule + semantic label-drop schedule &
Evaluates the final training formulation by adding dynamic scheduling and semantic dropout to the multi-scale hybrid model \\
\hline
\end{tabular}
\end{table*}

\subsubsection{Ablation Study to Evaluate Multi-Channel}
To evaluate the impact of the multi-channel conditioning design in the proposed framework, we conducted an ablation study using 201 validation images at a resolution of 768×768. The results are summarized in Table~\ref{tab:ablation_study}. The full model using eight input channels achieved the best overall performance with an SSIM of 0.9352, a PSNR of 26.95 dB, and an LPIPS of 0.1846. The results show that structural channels (edge and grayscale) are essential for high-quality crowd synthesis. Removing either the edge or grayscale channel caused degradation in image quality, with SSIM decreasing from 0.935 to approximately 0.117, PSNR falling to 6.82 dB, and LPIPS increasing sharply to above 0.70. The worst result occurred with edge-only conditioning, which corresponded to the standard P2P setup and resulted in a significant decrease in the image structure and texture. These findings confirm that both edge boundaries and grayscale intensity are essential for generating visually plausible crowd images consistent with structural conditioning inputs.

\noindent\begin{minipage}{\columnwidth}
\centering
\captionof{table}{Ablation study: eight-channel conditioning contributions.}
\label{tab:ablation_study}
\footnotesize
\setlength{\tabcolsep}{3pt}
\renewcommand{\arraystretch}{1.15}
\setlength{\arrayrulewidth}{0.6pt}
\setlength{\doublerulesep}{0pt}

\begin{tabular}{|l|c|c|c|c|l|}
\hline
Configuration$^{*}$ & SSIM & PSNR & LPIPS & $\Delta$SSIM & Impact\\
\hline
Full (E+G+D+T) & 0.935 & 26.95 & 0.185 & --- & Our baseline \\
\hline
No edges       & 0.117 & 6.82  & 0.700 & $-87.5\%$ & Catastrophic \\
\hline
No grayscale   & 0.117 & 6.82  & 0.707 & $-87.5\%$ & Catastrophic \\
\hline
Edges only     & 0.111 & 6.77  & 0.704 & $-88.1\%$ & Catastrophic \\
\hline
E+G only       & 0.934 & 26.77 & 0.188 & $-0.1\%$  & Minimal \\
\hline
No density     & 0.924 & 25.86 & 0.203 & $-1.2\%$  & Moderate \\
\hline
No temporal    & 0.913 & 24.27 & 0.199 & $-2.4\%$  & Moderate \\
\hline
\end{tabular}

\vspace{2pt}
\par\raggedright\footnotesize
$^{*}$E: edges, G: grayscale, D: density, T: temporal. $\Delta$SSIM denotes the relative change from the full model.
\end{minipage}

\vspace{4pt}

In contrast, removing contextual channels (density and temporal labels) resulted in a moderate reduction in performance. For example, excluding density labels decreased SSIM by only 1.2\%, while removing temporal labels reduced SSIM by 2.4\%. This indicates that the structural channels form the core representation required for high-fidelity synthesis. These attributes primarily enhance visual realism and attribute-specific appearance variation. The full eight-channel configuration achieved the highest performance in all metrics. However, using only edge and grayscale channels preserved 99.9\% of the SSIM, making it a minimally sufficient input for high-quality Hajj crowd image synthesis. Attribute channels, such as density and temporal labels, add moderate improvements in realism and provide better modulation over visual characteristics.

\subsubsection{Progressive Component-Integration Analysis: Ablation and Methodological Insights}
A staged integration analysis was conducted to examine how each architectural and training component contributed to P2P-H under limited-data, self-paired conditioning. The corresponding experimental design and quantitative results are summarized in Tables~\ref{tab:progressive_integration} and ~\ref{tab:e0e5_quant_test}, respectively. Table~\ref{tab:progressive_integration} summarizes the role of each architectural and training component, whereas Table ~\ref{tab:e0e5_quant_test} quantifies their effects on both distribution-level realism and pixel-level fidelity.

The results show that the dominant gain does not arise from loss-level refinement alone but from the introduction of hybrid structural-context conditioning. In particular, the transition 
from E1 to E2 produced a large improvement, with FID dropping from 517.0 to 73.17, PSNR increasing from 13.02 to 30.75, and SSIM increasing from 0.3435 to 0.9666. This change indicates that the conditioning design is the main source of representational gain in the proposed framework. E3 then further refined performance by adding stabilization-oriented components, including perceptual loss, feature matching, spectral normalization, R1 regularization, and EMA, yielding the strongest single-discriminator configuration with FID = 50.32 and the highest pixel-level fidelity in the study (PSNR = 32.17, SSIM = 0.9705). 

A second important finding appeared in E4. When multi-scale discrimination and image-only DiffAug were introduced simultaneously, performance degraded rather than improved. FID increased from 50.32 to 86.34, KID increased from 0.0044 to 0.0318, PSNR dropped from 32.17 to 27.92, and LPIPS increased from 0.1163 to 0.1687. This result suggests that in a limited-data, self-paired setting, stronger adversarial pressure and stronger regularization are not automatically beneficial when applied together. The degradation was recovered in E5, where dynamic loss scheduling and semantic label-drop regularization restored stability and produced the best distribution-level realism in the study, with FID = 48.01 and KID = 0.0023, while still maintaining strong fidelity (PSNR = 30.37, SSIM = 0.9610). Taken together, Tables~\ref{tab:progressive_integration} and ~\ref{tab:e0e5_quant_test} show that the contribution of P2P-H cannot be explained by the independent addition of borrowed components alone. Rather, the results suggest a practical design insight for this setting: hybrid structural-context conditioning provides the principal gain, stabilization improves perceptual quality, and multi-scale adversarial training becomes effective only when paired with explicit training-time control.

\begin{table}[!htbp]
\caption{Quantitative results of the E0--E5 progressive integration study on the test set.}
\label{tab:e0e5_quant_test}
\centering

\footnotesize
\setlength{\tabcolsep}{3pt} 
\renewcommand{\arraystretch}{1.15}
\setlength{\arrayrulewidth}{0.4pt}

\begin{tabular*}{\columnwidth}{@{\extracolsep{\fill}}|c|c|c|c|c|c|c|}
\hline
\textbf{Metric} & \textbf{E0} & \textbf{E1} & \textbf{E2} & \textbf{E3} & \textbf{E4} & \textbf{E5} \\
\hline
\textbf{FID $\downarrow$}   & 570.9  & 517.0  & 73.17  & 50.32  & 86.34  & 48.01 \\
\hline
\textbf{KID $\downarrow$}   & 0.5210 & 0.4112 & 0.0423 & 0.0044 & 0.0318 & 0.0023 \\
\hline
\textbf{SSIM $\uparrow$}    & 0.4011 & 0.3435 & 0.9666 & 0.9705 & 0.9534 & 0.9610 \\
\hline
\textbf{PSNR $\uparrow$}    & 14.00  & 13.02  & 30.75  & 32.17  & 27.92  & 30.37 \\
\hline
\textbf{LPIPS $\downarrow$} & 0.5501 & 0.6517 & 0.1542 & 0.1163 & 0.1687 & 0.1332 \\
\hline
\end{tabular*}
\end{table}

\FloatBarrier

\subsubsection{Semantic Label-Sweep Controllability Analysis Under Fixed Structural Conditioning}
\begin{figure}[!htbp]
\centering
\includegraphics[width=\columnwidth,height=0.36\textheight]{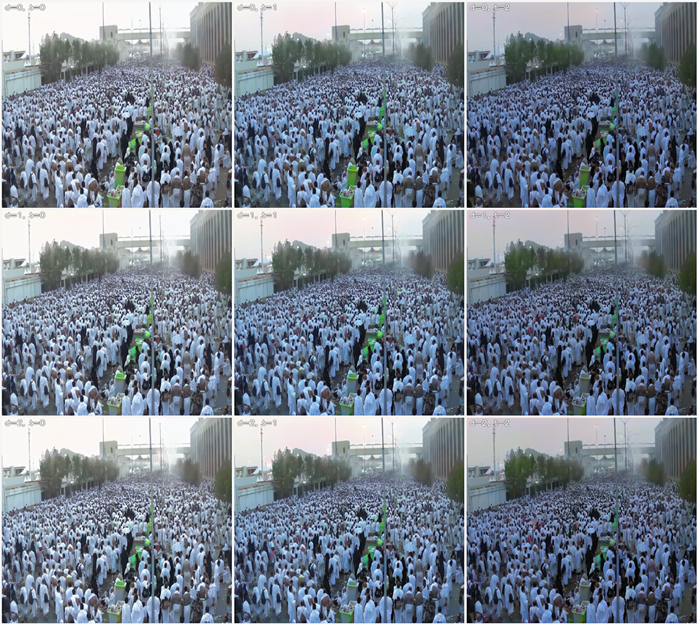}
\caption{Semantic label-sweep analysis under fixed structural conditioning. The edge and grayscale channels are held constant for the same scene, while the time-of-day label $t\in\{0,1,2\}$ is varied across columns and the density label $d\in\{0,1,2\}$ is varied across rows. Time-of-day conditioning produces clear photometric modulation, whereas density effects remain limited because the fixed edge and grayscale channels constrain the crowd geometry.
}
\label{fig:control}
\end{figure}

To evaluate how much variability and controllability P2P-H can achieve under different semantic conditions, we performed a controlled semantic label-sweep experiment while keeping the structural conditioning fixed. For each selected scene, the edge map and grayscale image were held constant, while the one-hot labels of the time-of-day attribute ($t\in\{0,1,2\}$)and density class ($d\in\{0,1,2\}$) were systematically varied. This protocol produced multiple outputs from the same structural input and therefore evaluated whether the generator responds to semantic conditioning beyond deterministic reconstruction. Representative results are shown in Fig.~\ref{fig:control}.

Time-of-day modulation: Across the time-of-day columns (t = 0 →1→2), the generated outputs exhibited consistent photometric changes, including variations in brightness, contrast, color temperature, global illumination, background tone, and shadow appearance. Because the edge and grayscale channels are fixed, the dominant scene layout remains stable. These results indicate that the time-of-day attribute was learnable and can be expressed mainly as appearance-level modulation under fixed structural constraints.

Density-label effects under strong structural constraints: Across density rows (d = 0 →1→2), the visual changes were less pronounced. This behavior was expected in the proposed self-paired formulation because the edge and grayscale channels preserve the dominant crowd geometry. Thus, large geometric changes, such as producing a clearly different number or arrangement of visible pedestrians, were constrained by the fixed structural input. Therefore, the density label is interpreted as a weak semantic cue that may influence local texture, contrast, and crowd-appearance tendencies, rather than as an independent mechanism for controlling the exact number of visible individuals.

Implication for controllability: The label-sweep results show that P2P-H supports limited semantic controllability under fixed structural conditioning. The controllability is asymmetric: time-of-day labels produce clearer photometric modulation, whereas density labels have weaker effects because the structural channels dominated the generated layout. Thus, Fig.~\ref{fig:control} provides evidence of appearance-level semantic modulation, rather than evidence of fully unconstrained crowd-layout generation.

To further quantify semantic controllability under fixed structural conditioning, we evaluated the semantic label-sweep outputs using 553 complete scenes. For each scene, the edge and grayscale channels were fixed, while the density label \(d \in \{0,1,2\}\) and the time-of-day label \(t \in \{0,1,2\}\) were varied, producing nine outputs per scene. As reported in Table~\ref{tab:semantic_controllability}, changing the time-of-day label produced a mean luma difference of \(\Delta \mu_Y(t=0,t=2)=0.17165 \pm 0.02228\), together with \(D^{\mathrm{time}}_{\mathrm{LPIPS}}=0.05760 \pm 0.02684\). This indicates measurable photometric and appearance-level controllability. In contrast, varying the density label produced \(D^{\mathrm{density}}_{\mathrm{LPIPS}}=0.02886 \pm 0.01738\) and an edge-density change of \(\Delta \rho_e(d=2,d=0)=-0.00080 \pm 0.00128\),indicating that density controllability is weaker under fixed structural constraints. The pairwise Edge IoU of \(0.89986 \pm 0.03410\) further shows that the dominant scene structure is largely preserved during semantic label manipulation. These findings support the interpretation that P2P-H provides stronger controllability over photometric attributes than over true crowd geometry, which is expected in the self-paired formulation where the edge and grayscale channels constrain the layout.

\noindent
\begin{table}[!htbp]
\caption{Quantitative semantic controllability under fixed structural conditioning.}
\label{tab:semantic_controllability}
\centering
\scriptsize
\begin{tabular}{|l|l|c|}
\hline
\textbf{Factor} & \textbf{Metric} & \textbf{Result} \\
\hline
Time-of-day & $\Delta \mu_Y(t=0,t=2)$ & $0.17165 \pm 0.02228$ \\
\hline
Time-of-day & $D^{\mathrm{time}}_{\mathrm{LPIPS}}$ & $0.05760 \pm 0.02684$ \\
\hline
Density & $D^{\mathrm{density}}_{\mathrm{LPIPS}}$ & $0.02886 \pm 0.01738$ \\
\hline
Density & $\Delta \rho_e(d=2,d=0)$ & $-0.00080 \pm 0.00128$ \\
\hline
Structure & Pairwise Edge IoU & $0.89986 \pm 0.03410$ \\
\hline
\end{tabular}
\end{table}

\subsubsection{Latent-Noise Diversity Analysis of P2P-H-Z Under Fixed Structural and Semantic Conditioning}

To evaluate whether P2P-H can produce non-identical outputs under identical structural and semantic conditions, we implemented a latent-noise variant, denoted P2P-H-Z. In this variant, the edge, grayscale, density, and time-of-day conditioning channels were held fixed, while latent noise vectors were injected into multiple decoder stages of the generator. For each held-out validation condition, multiple outputs were generated using different latent samples and pairwise LPIPS was then computed between outputs generated from the same conditioning tensor to quantify stochastic diversity under fixed conditioning.

The noise-augmented variant achieved a pairwise LPIPS diversity of \(0.0248 \pm 0.0094\) on the held-out validation set, demonstrating measurable stochastic variability relative to a deterministic mapping. At the same time, it maintained conditional synthesis quality, with FID \(=50.08\), SSIM \(=0.9580\), PSNR \(=27.94\) dB, and LPIPS-to-real \(=0.1422\) on the complete validation set. The results are summarized in Table~\ref{tab:stochastic_diversity}. These results show that latent noise can introduce non-identical outputs under fixed structural and semantic conditioning while maintaining reasonable conditional fidelity.

\begin{table}[!htbp]
\caption{Stochastic diversity analysis of P2P-H-Z under fixed conditioning.}
\label{tab:stochastic_diversity}
\centering
\resizebox{\columnwidth}{!}{%
\begin{tabular}{|l|c|c|c|c|c|c|}
\hline
\textbf{Variant} & \textbf{Noise} & \textbf{Pairwise LPIPS} & \textbf{FID} & \textbf{SSIM} & \textbf{PSNR (dB)} & \textbf{LPIPS-real} \\
\hline
P2P-H-Z & Yes & \(0.0248 \pm 0.0094\) & \(50.08\) & \(0.9580\) & \(27.94\) & \(0.1422\) \\
\hline
\end{tabular}%
}
\end{table}

The induced variation was primarily appearance-level,
including local texture, tonal variation, contrast differences, and minor illumination changes, while the dominant scene structure remained largely preserved by the edge and grayscale conditioning channels. Therefore, this experiment shows that stochastic variation can be incorporated into P2P-H under fixed conditioning; however, it should not be interpreted as evidence of fully unconstrained crowd-scene generation. Instead, these results support the positioning of P2P-H as a structure-guided conditional synthesis framework with limited appearance-level stochasticity.

\begin{figure}[!htbp]
\centering
\includegraphics[width=\columnwidth]{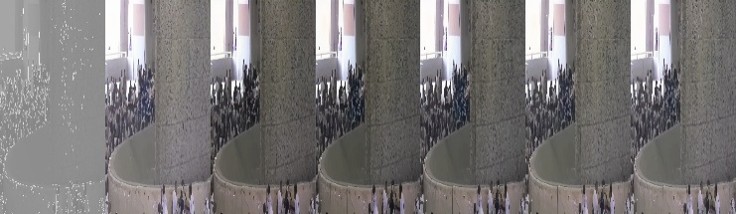}
\caption{Fixed-condition multi-sample outputs from the latent-noise P2P-H-Z variant. For each row, the structural conditioning is held fixed, while multiple outputs are generated using different latent samples. The results show limited appearance-level variation in texture, tone, contrast, and illumination, while the dominant scene structure remains largely preserved.
}
\label{fig:noise}
\end{figure}

The visual examples in Fig.~\ref{fig:noise} further illustrate that different latent samples produce non-identical outputs under the same structural and semantic conditions. However, the induced variation is primarily appearance-level, including local texture, tonal variation, contrast differences, and minor illumination changes, while the dominant scene structure remains largely preserved by the edge and grayscale conditioning channels. Therefore, this experiment shows that stochastic variation can be incorporated into P2P-H under fixed conditioning. The result supports the positioning of P2P-H as a structure-guided conditional synthesis framework with limited appearance-level stochasticity.

Taken together, the semantic label-sweep analysis and the noise-augmented variant provide complementary evidence regarding diversity under the self-paired formulation. The label-sweep experiment shows that P2P-H responds more strongly to time-of-day conditioning than to density conditioning under fixed edge and grayscale inputs. The noise-augmented variant further shows that non-identical outputs can be produced from the same structural and semantic condition. However, both analyses indicate that the induced variation is primarily appearance-level, while the dominant crowd layout remains constrained by the structural channels. Therefore, the proposed framework is best interpreted as structure-guided conditional synthesis with limited semantic and stochastic appearance variation, rather than fully unconstrained crowd-scene generation.

\subsubsection{Cross-Dataset Generalization}

The results of the quantitative evaluation conducted using the testing sets are collected in Table~\ref{tab:cross_domain}. 

\begin{table}[!htbp]
\centering
\caption{Cross-domain evaluation on UCF\_CC\_50 and UCF-QNRF.}
\label{tab:cross_domain}
\footnotesize
\setlength{\tabcolsep}{3pt}
\renewcommand{\arraystretch}{1.15}
\setlength{\arrayrulewidth}{0.4pt}

\begin{tabular}{|l|c|c|c|c|c|}
\hline
Dataset & FID $\downarrow$ & SSIM $\uparrow$ & PSNR (dB) $\uparrow$ & LPIPS $\downarrow$ & KID $\downarrow$ \\
\hline
UCF\_CC\_50 ($N{=}50$) & 3.89 & 0.982 & 35.21 & 0.049 & 0.0326 \\
\hline
UCF-QNRF ($N{=}334$) & 61.90 & 0.882 & 22.27 & 0.261 & 0.0090 \\
\hline
\end{tabular}
\end{table}

The proposed model achieved strong reconstruction-oriented image quality under cross-dataset structural conditioning, obtaining FID = 3.89, SSIM = 0.982, and PSNR = 35.21 dB on UCF\_CC\_50, and FID = 61.90, SSIM = 0.882, and PSNR = 22.27 dB on UCF-QNRF. These results indicate that the learned conditional mapping transfers to structurally derived inputs from unseen crowd datasets. In particular, the generated outputs preserved the dominant scene geometry and density layout even when viewpoint, crowd scale, and illumination differed from the Hajj training data. When comparing the results on UCF-QNRF with those on UCF\_CC\_50 dataset, the higher LPIPS score on UCF-QNRF reflects some color desaturation due to differing illumination statistics, while FID/KID values confirm robust distributional similarity across unseen datasets. These experiments support the ability of P2P-H to perform cross-dataset conditional reconstruction under unseen structural inputs. These qualitative examples are consistent with the quantitative cross-domain results reported in Table~\ref{tab:cross_domain}. However, in this self-paired setting, these results are interpreted as evidence of cross-dataset conditional reconstruction capability.

Figs.~\ref{fig:test_ucf_qnrf}, \ref{fig:test_ucf_cc50}, and \ref{fig:test_hajjv2} present qualitative results of the proposed P2P-H framework on the in-domain HAJJv2 dataset and the unseen UCF-QNRF and UCF\_CC\_50 datasets. The generated images preserved the dominant scene structure and produced visually apparent crowd textures under different illumination, viewpoint, and density conditions. These qualitative examples are consistent with the quantitative cross-domain results reported in Table~\ref{tab:cross_domain}. However, in this self-paired


\FloatBarrier 

\begin{strip}
\centering
\setlength{\abovecaptionskip}{2pt}
\setlength{\belowcaptionskip}{2pt}

\includegraphics[width=\textwidth,height=0.29\textheight]{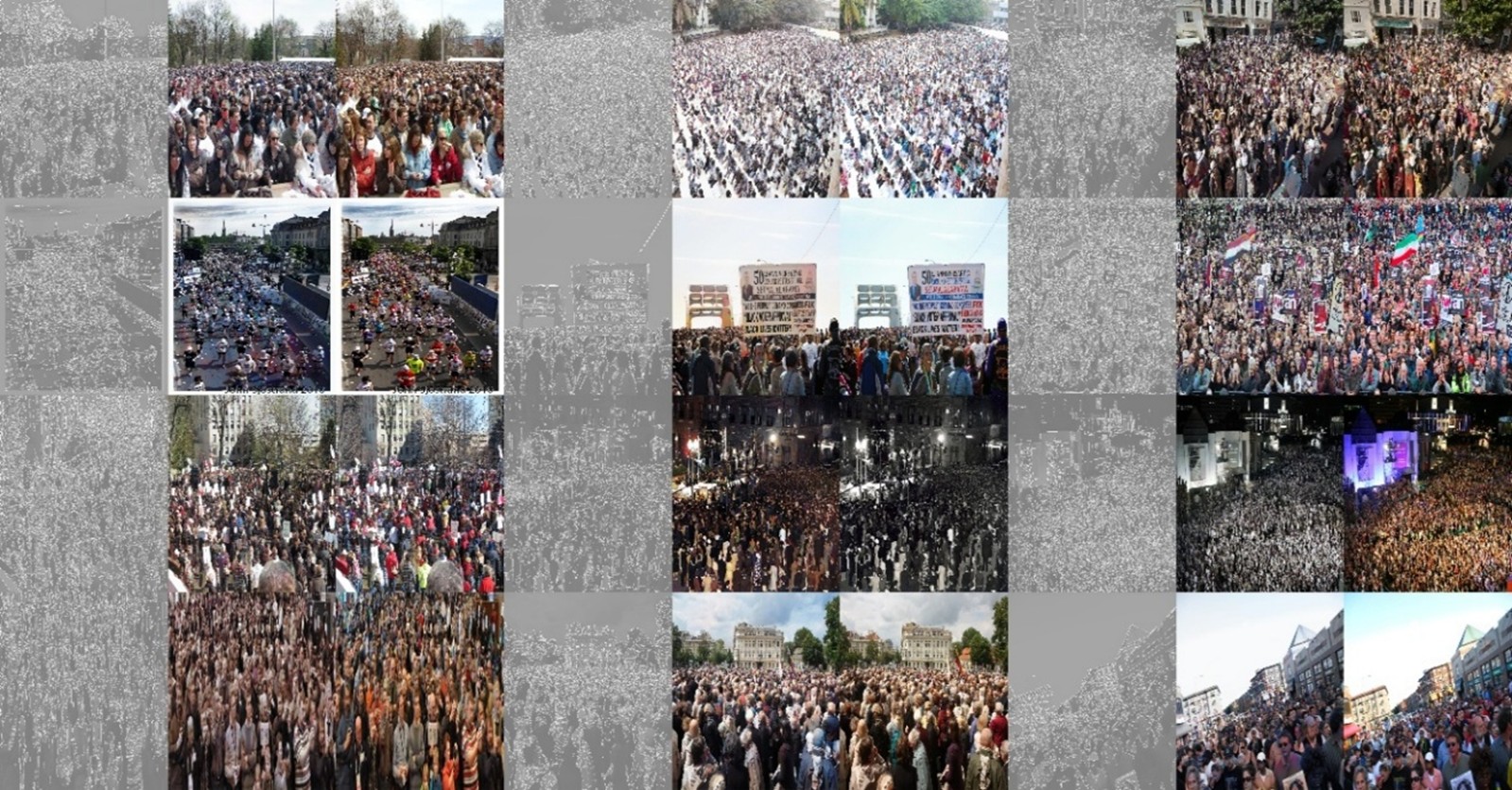}
\captionof{figure}{Testing P2P-H on the UCF\_QNRF crowd counting dataset. The left column displays the grayscale edge map input, the center column shows the synthetic image generated by P2P-H, and the right column presents the corresponding real image.}
\label{fig:test_ucf_qnrf}

\vspace{4pt}

\includegraphics[width=\textwidth,height=0.29\textheight]{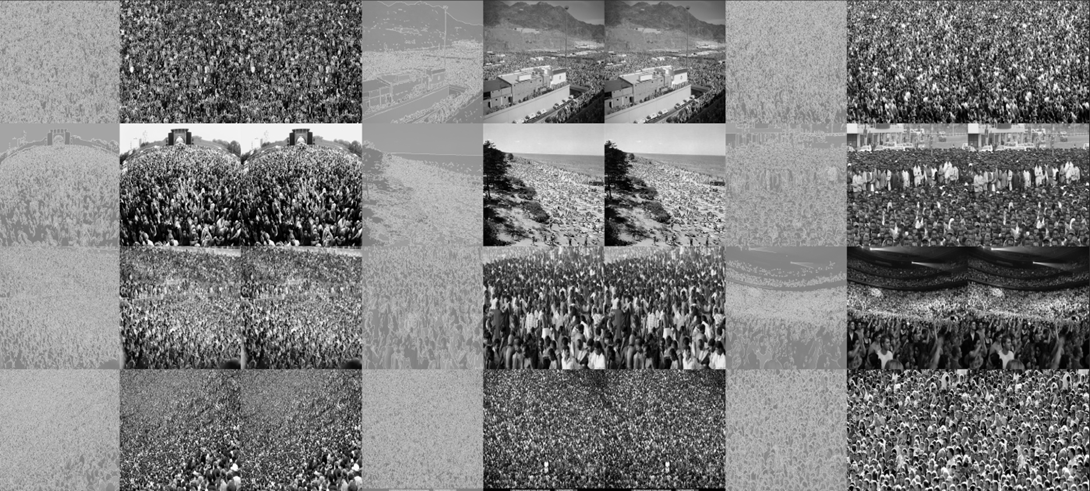}
\captionof{figure}{Testing P2P-H on the UCF\_CC\_50 crowd counting dataset. The left column displays the grayscale edge map input, the center column shows the synthetic image generated by P2P-H, and the right column presents the corresponding real image.}
\label{fig:test_ucf_cc50}

\vspace{4pt}

\includegraphics[width=\textwidth,height=0.29\textheight]{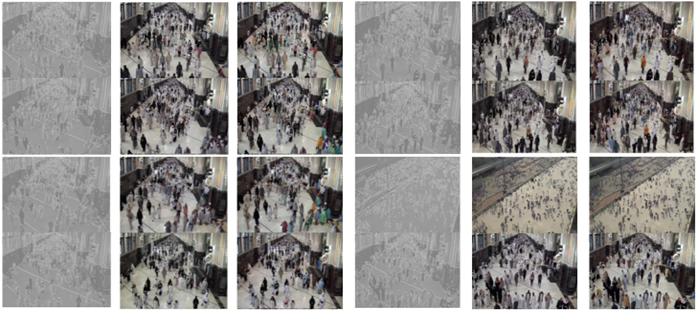}
\captionof{figure}{Testing P2P-H on the HAJJv2 crowd dataset. The left column displays the grayscale edge map input, the center column shows the synthetic image generated by P2P-H, and the right column presents the corresponding real image.}
\label{fig:test_hajjv2}

\end{strip}

 \FloatBarrier
 
setting, these results are interpreted as evidence of cross-dataset conditional reconstruction capability.

\subsubsection{Training Dynamics}
\begin{figure}[!htbp]
\centering
\includegraphics[width=\columnwidth,height=0.55\textheight,keepaspectratio]{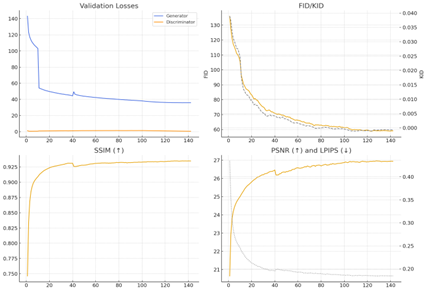}
\caption{Validation performance curves for the proposed P2P-H model.}
\label{fig:GPU}
\end{figure}

Fig.~\ref{fig:GPU} illustrates the validation dynamics of the proposed P2P-H model. The generator loss decreased smoothly and stabilized around epoch 110. The discriminator loss remained low and flat throughout training. This pattern indicates balanced adversarial training. The image quality consistently improved. The values of FID and KID decreased over time, while the SSIM and PSNR values steadily increased. The LPIPS score dropped to approximately 0.20. Overall, these patterns support the effectiveness of the training approach. These trends validate the stability and generalization capacity of P2P-H in generating high-quality attribute-conditioned hajj crowd images without overfitting. These trends show improved alignment with the ground truth, and the training appears to be stable. All curves progressed smoothly and no overfitting was observed. The model generated high-quality structure-guided crowd images consistent with the conditioning inputs.

\subsubsection{Computational Efficiency}

All experiments were conducted with Google Colab Pro. We used a single NVIDIA A100-SXM4 GPU with 40 GB VRAM. The proposed P2P-H framework required approximately 1.5 hours for complete training, while the standard P2P baseline required 2 hours for training and SGAN2-ADA required nearly 48 hours for training. These results indicate the practical efficiency of P2P-H in balancing training speed and output quality for large-scale crowd image synthesis.
\FloatBarrier

\subsection{EVALUATE GENERATED DATA FROM PROPOSED P2P-H MODEL}
The output quality of the deployed model was verified through two independent assessments.

\subsubsection{Feature-Space Distribution Analysis via t-SNE Visualization}
To qualitatively evaluate feature-space organization between real and synthetic Hajj crowd images, deep feature embeddings were extracted using the pretrained InceptionV3 model. Each image was resized to 299×299pixels and processed through the global average pooling layer to obtain 2048-dimensional feature vectors. PCA was applied before t-SNE using 50 components. The perplexity parameter was set to 30, and a fixed random seed was used to project the high-dimensional features onto a two-dimensional manifold for visualization. These visualizations were generated using 993 real images from the training set and 4,982 synthetic images from the proposed P2P-H model.

\begin{figure}[!htbp]
\centering
\includegraphics[width=\columnwidth,height=0.20\textheight]{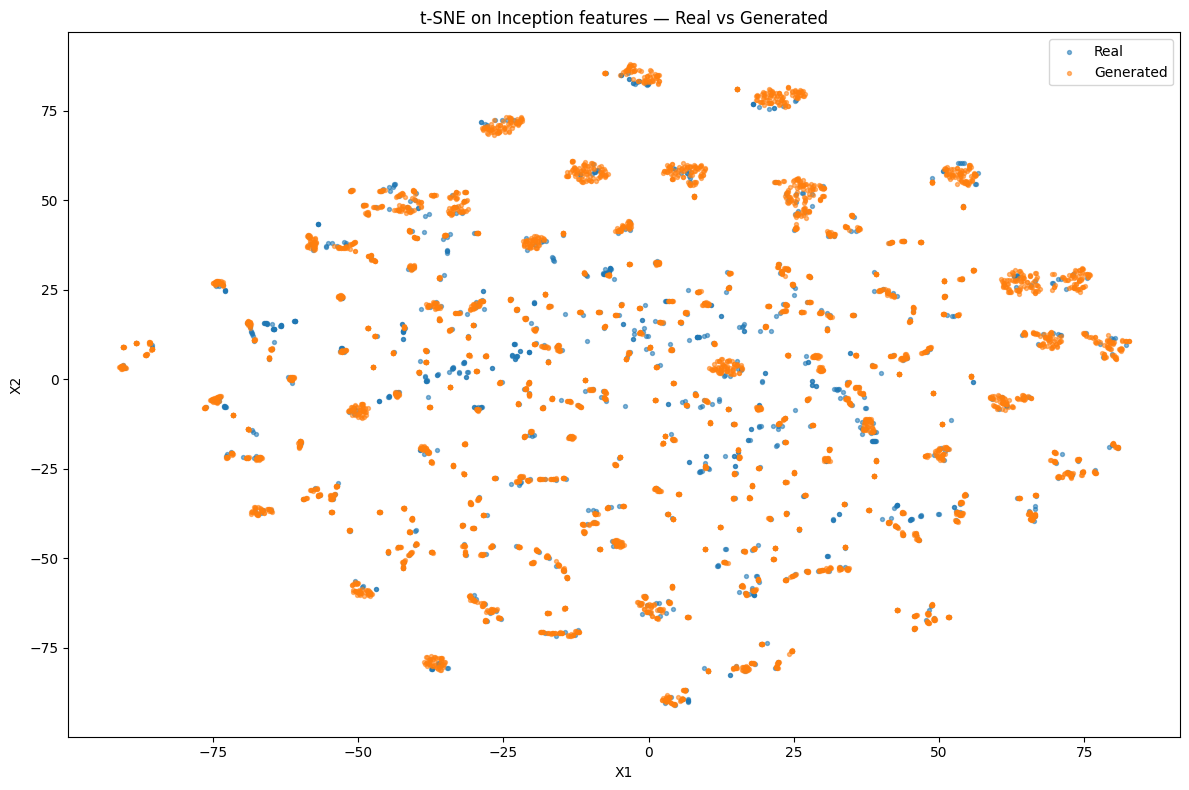}
\caption{t-SNE visualization of 4.982 synthetic images (orange) compared with all the real images from the training set (blue).}
\label{fig:control-t-SNE}
\end{figure}

Real vs. generated feature-space organization: Fig.~\ref{fig:control-t-SNE} presents the t-SNE plot of real images and generated samples. The overlap between the real and generated samples provides qualitative evidence of feature-space similarity under the selected InceptionV3 embedding protocol. However, this visualization should not be interpreted as definitive proof that the real and generated distributions are identical. Some peripheral gaps remain, which may be associated with rare viewpoints, sparse crowd cases, or underrepresented visual conditions.

\begin{figure}[!htbp]
\centering
\includegraphics[width=\columnwidth,height=0.20\textheight]{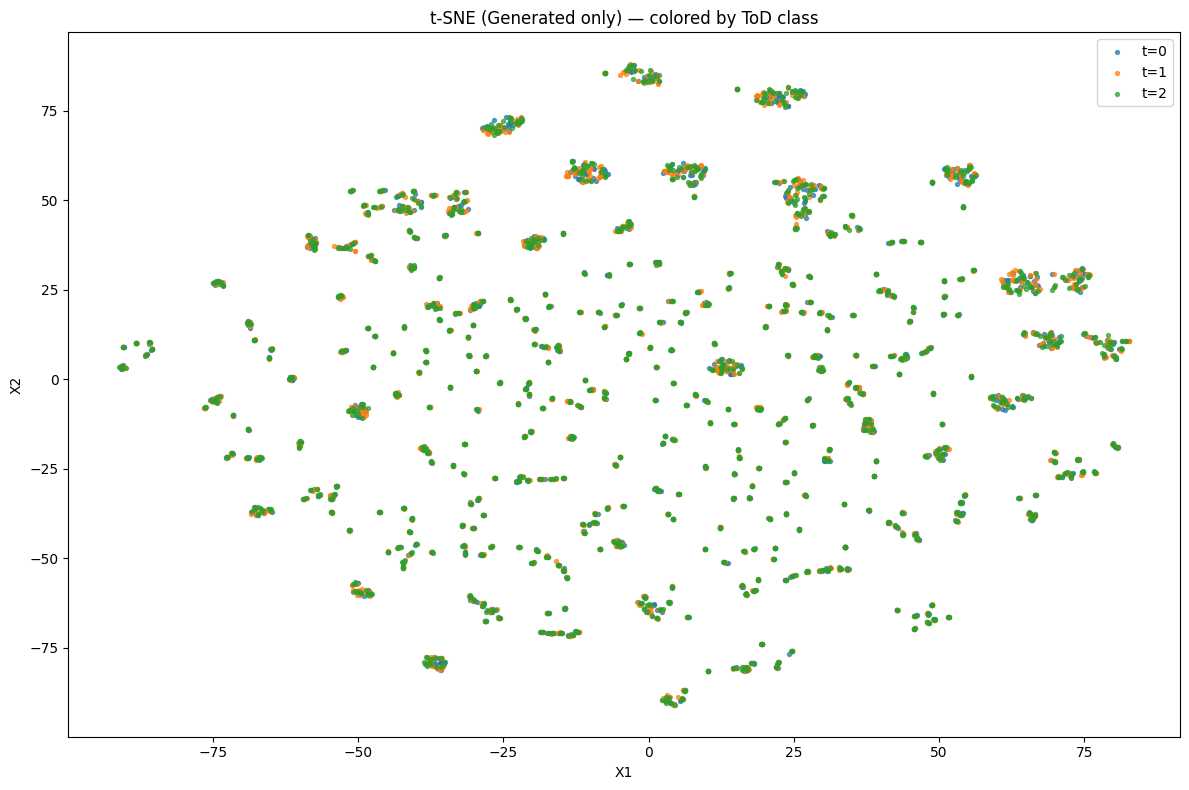}
\caption{t-SNE visualization of generated samples from the proposed P2P-H model. Colored by time-of-day class (t = 0, 1, 2).}
\label{fig:T-SNE-T}
\end{figure}

Interpretability via time-of-day conditioning: Fig.~\ref{fig:T-SNE-T} shows the generated samples color-coded by time-of-day (ToD) class, where (t=0), (t=1), and (t=2) correspond to bright, moderate, and dim illumination conditions, respectively. The visualization shows partial local organization according to the time-of-day labels, although the classes are not fully separated globally. This observation is consistent with the time-of-day classification accuracy of 74.2\%, suggesting that P2P-H is moderately responsive to illumination-related conditioning under this evaluation protocol.

\FloatBarrier

\begin{figure}[!htbp]
\centering
\includegraphics[width=\columnwidth,height=0.20\textheight]{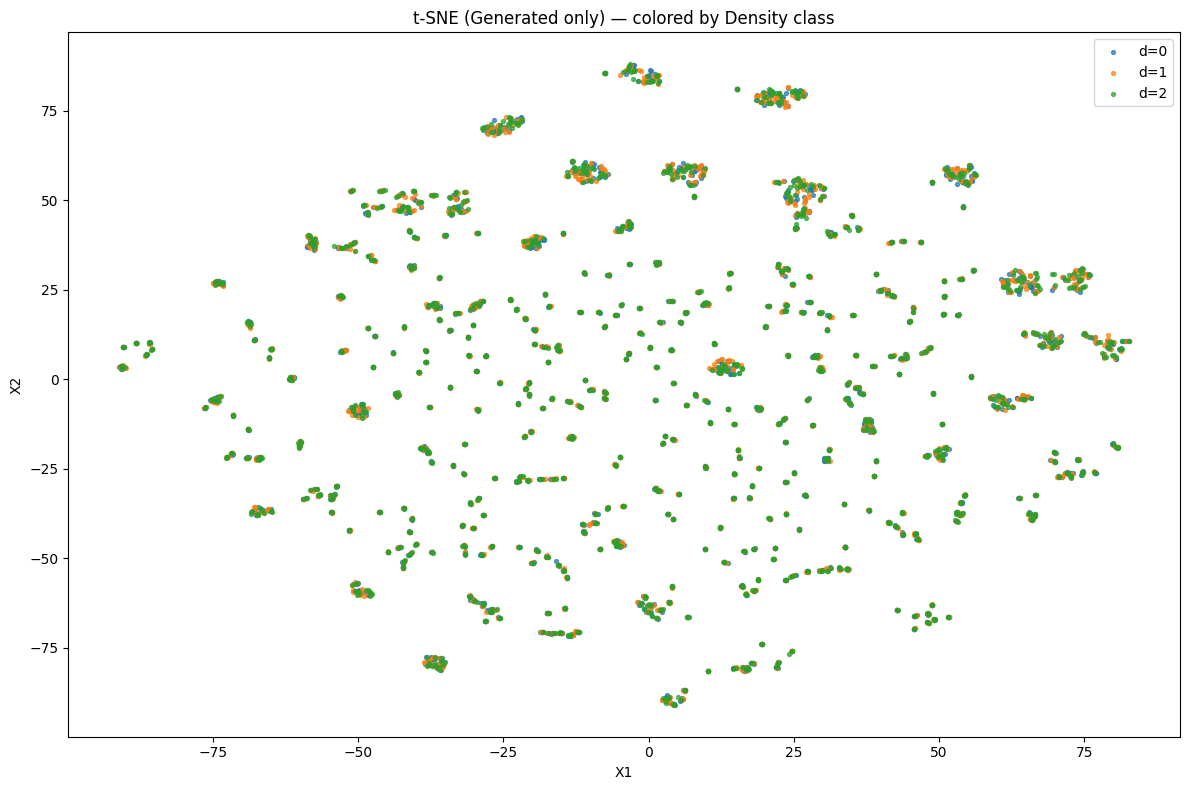}
\caption{t-SNE visualization of generated samples from the proposed P2P-H model. Colored by crowd density class (d = 0, 1, 2).}
\label{fig:t-SNE-D}
\end{figure}

Crowd density class organization: Fig.~\ref{fig:t-SNE-D} presents the generated samples color-coded by density class, where (d=0), (d=1), and (d=2) correspond to low, medium, and high density levels, respectively. Compared with the time-of-day visualization, the density-labeled samples show stronger overlap and weaker class separation. This behavior is consistent with the density classification accuracy of 48.1\%, indicating that density-related conditioning is less separable in the generated feature space.

The t-SNE visualizations provide qualitative supporting evidence that P2P-H exhibits feature-space overlap between real and generated samples, interpretable organization related to weak semantic attributes, and partial sensitivity to crowd-relevant conditional inputs. The time-of-day labels show clearer local organization than the density labels, indicating stronger responsiveness to illumination-related attributes, whereas density control remains constrained by the fixed structural channels.

\FloatBarrier
\subsubsection{Binary Classifier Validation} 

\begin{figure}[!htbp]
\centering
\includegraphics[width=\columnwidth,height=0.20\textheight]{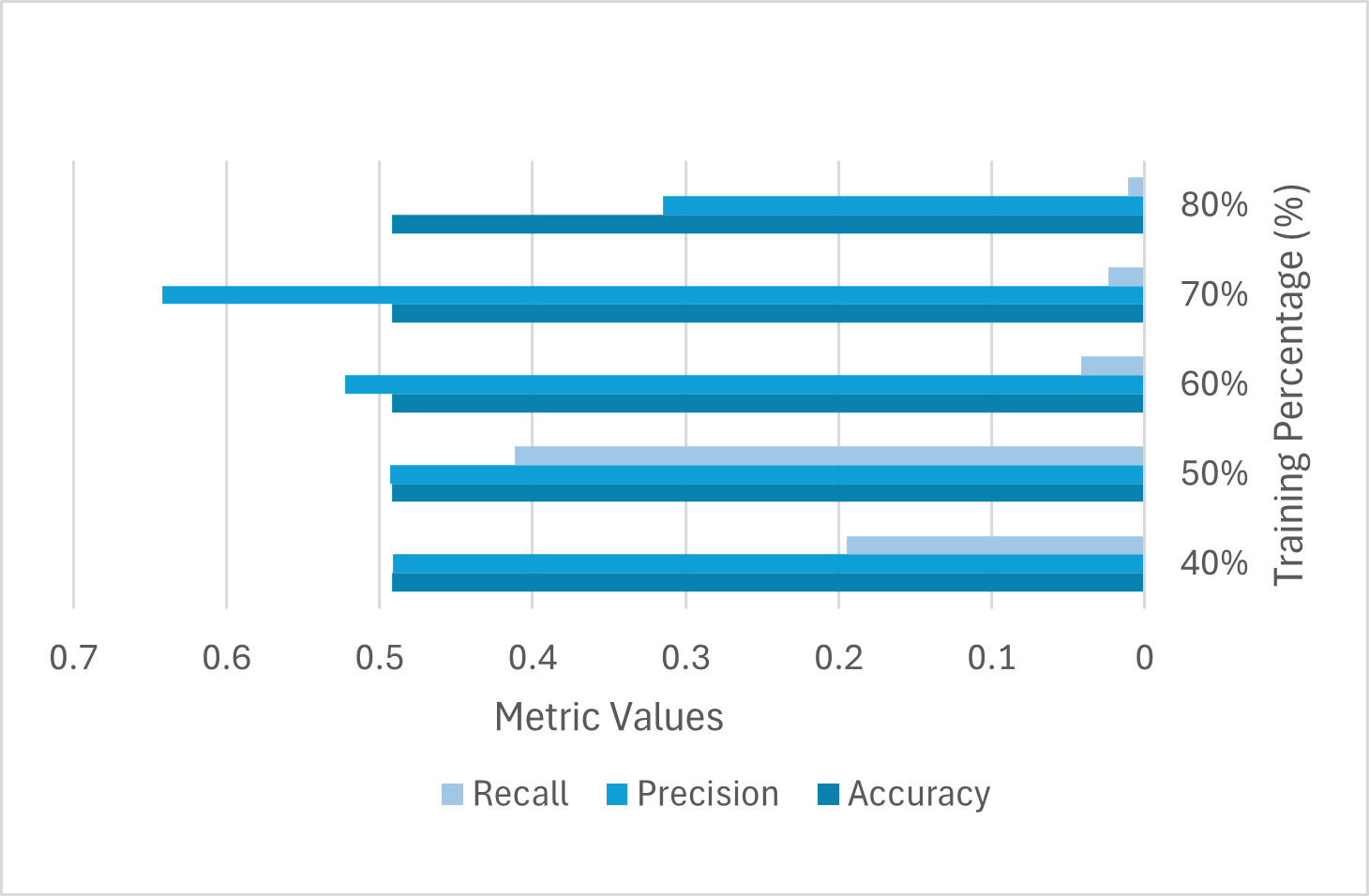}
\caption{Performance metrics of the binary classifier across different training percentages: 40\%, 50\%, 60\%,70\%,80\%.}
\label{fig:BC}
\end{figure}

We evaluated the generated images using a binary CNN classifier trained to distinguish real from synthetic Hajj crowd images. Real images were labeled as 0, and synthetic images were labeled as 1. A balanced dataset of 993 images was used, and five training/testing splits from 40\% to 80\% were evaluated, as shown in Fig.~\ref{fig:BC}.

The binary classifier results are interpreted as an auxiliary real–synthetic separability check. Across the evaluated configurations, accuracy ranged from 0.4914 to 0.5050, precision from 0.3333 to 0.6364, recall from 0.0151 to 0.4415, and F1-score from 0.0288 to 0.4645. The accuracy values remained close to random-chance performance (0.5)across all training splits, with the closest value observed at the 50\% training split (accuracy =0.4914). Under this specific classifier architecture, preprocessing setting, and train/test protocol, the classifier did not learn a reliable boundary between the selected real and synthetic samples.

\FloatBarrier

\subsubsection{Nearest-Neighbor Memorization Test}
\begin{figure}[!htbp]
\centering
\includegraphics[width=\columnwidth,height=0.20\textheight]{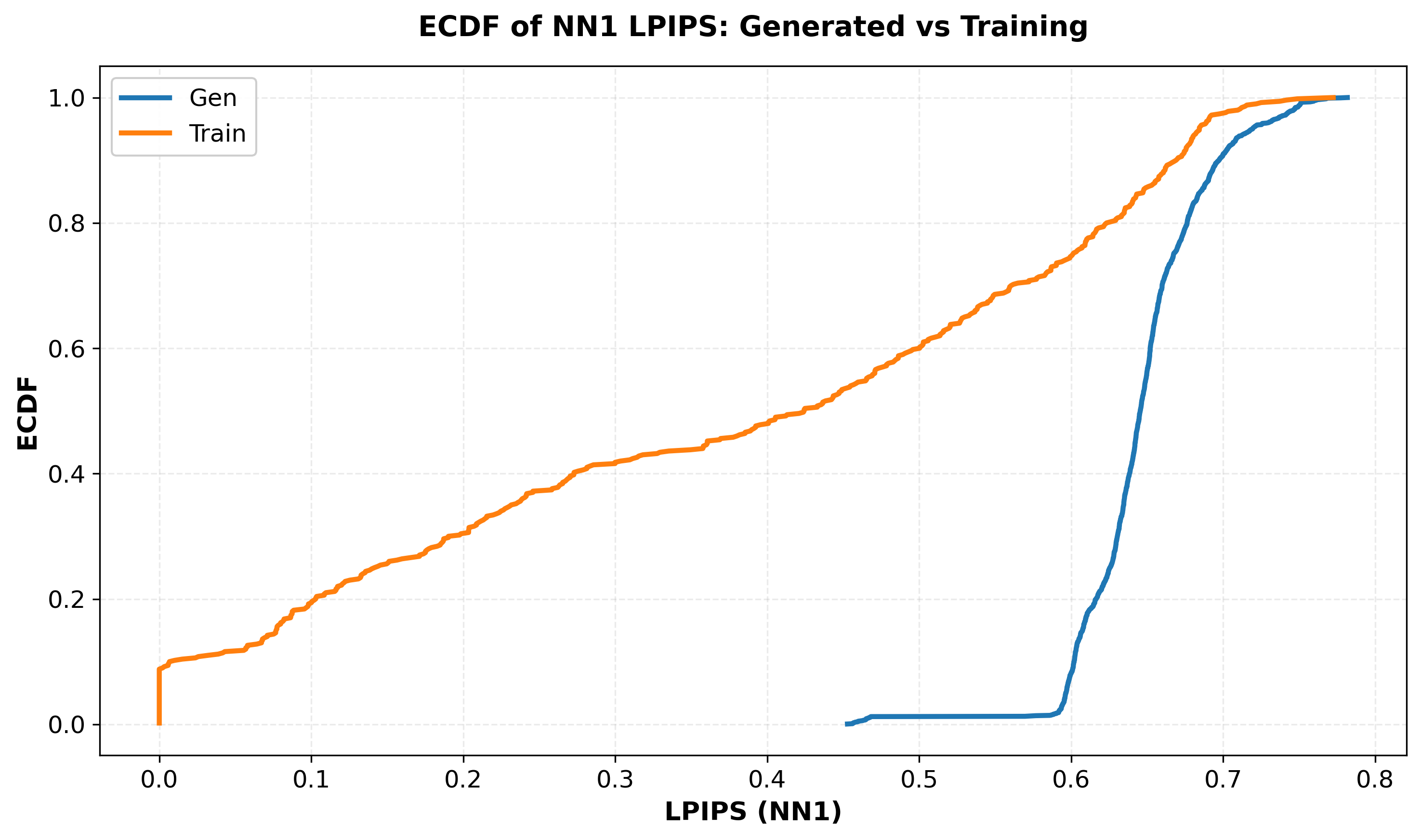}
\caption{ECDF of LPIPS to the nearest neighbor (NN1). The blue curve reports LPIPS between generated samples from HAJJv2 data and their closest training images retrieved in CLIP embedding space. The orange curve provides the Train→Train baseline, measuring LPIPS between each training image and  its nearest neighbor within the training set.}
\label{fig:ECDF-Hajjv2}
\end{figure}

\begin{figure}[!htbp]
\centering
\includegraphics[width=\columnwidth,height=0.20\textheight]{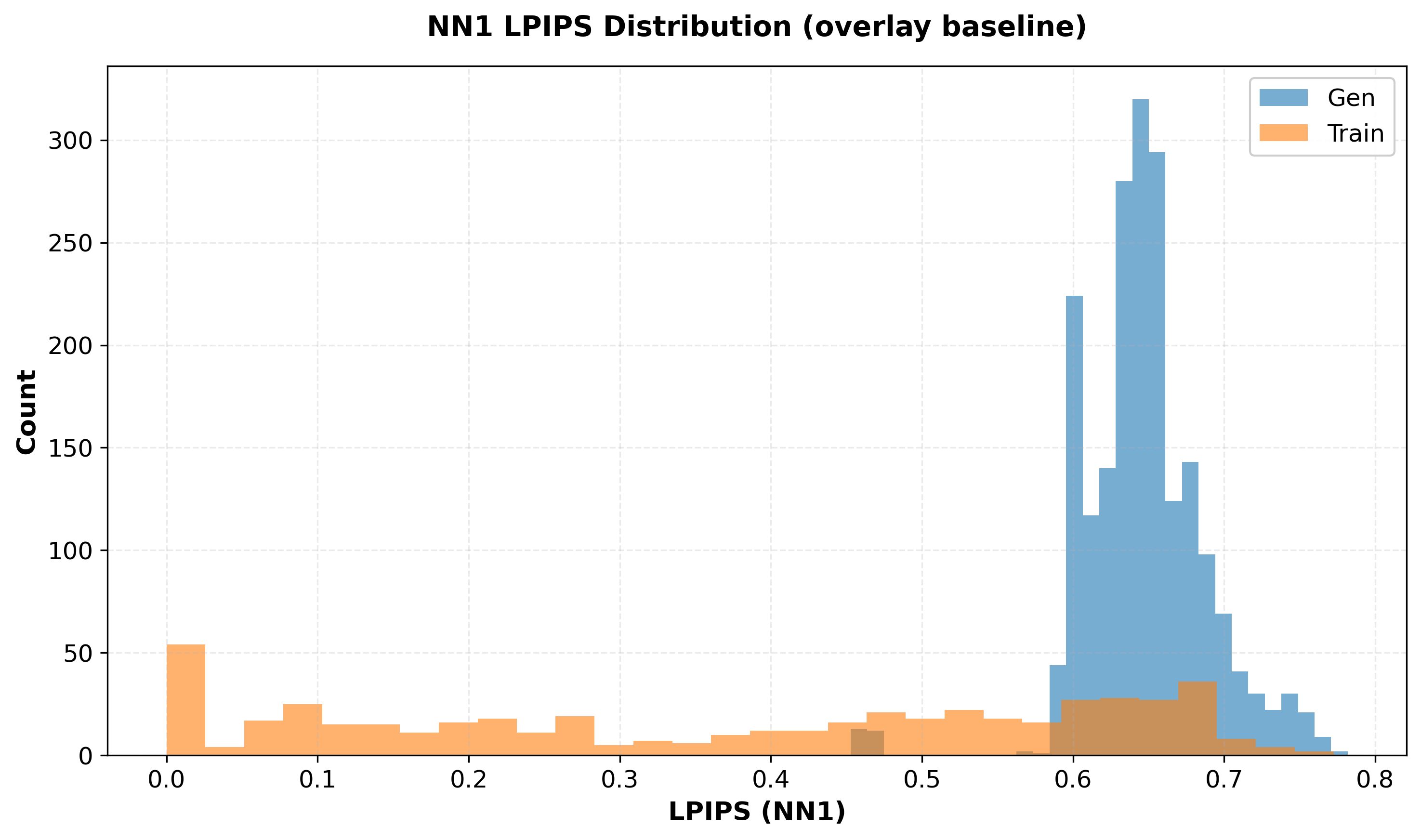}
\caption{Distribution overlay showing generated distances compared to training baseline. Histogram overlay of NN1 LPIPS distances comparing Generated against a Train baseline. The generated distribution is shifted toward higher LPIPS values, indicating no perceptual near-duplicate reproduction of training images.}
\label{fig:Hajjv2-histogram}
\end{figure}

To assess whether the generator memorizes or directly reproduces training examples, we conducted a nearest-neighbor analysis in feature space.For the in-domain nearest-neighbor analysis on the HAJJv2 dataset, we used the training data as the retrieval pool. We assessed the n = 2,039 generated output samples selected randomly from held-out HAJJv2 test input. The NN1 LPIPS (Generated→Train) distribution was 0.6474 ± 0.0418 (mean ± std), with min = 0.4529 and max = 0.8286 as summarized in Table 14. Train→Train NN1-LPIPS (VGG): mean ± std = 0.3562 ± 0.2217, (min = 0.112, max = 0.7180). In comparison to the Train→Train baseline exhibited much lower distances reflecting natural near-duplicate video frames. The two distributions were clearly different. As illustrated in by the ECDF in Fig.~\ref{fig:ECDF-Hajjv2} and the histogram overlays in Fig.~\ref{fig:Hajjv2-histogram}, the Generated→Train distribution had a right shift relative to the Train→Train distribution. These samples remained perceptually distinct from their closest retrieved training neighbors.

\begin{figure*}[!t]
\centering
\includegraphics[width=\textwidth,height=0.68\textheight]{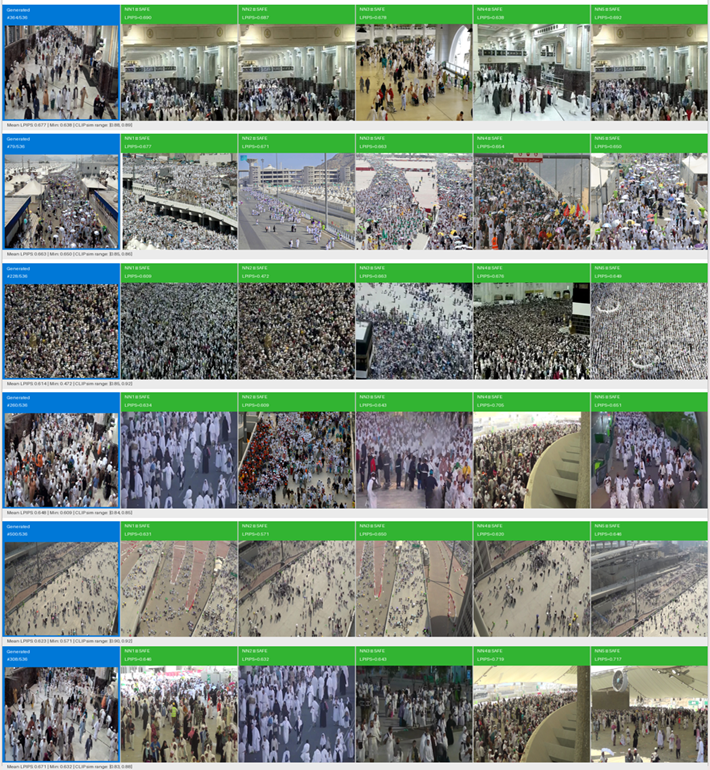}
\caption{Representative nearest-neighbor grids. Each row shows a generated image from HAJJv2 dataset followed by its top-5 nearest training images retrieved in CLIP embedding space (cosine similarity), with similarity annotations. NN1 denotes the closest neighbor among the top-5.}
\label{fig:NN-Hajjv2-22}
\end{figure*}

Although CLIP retrieval can return neighbors with high embedding similarity, the corresponding perceptual distances remain large. Even the most similar case retrieved  by CLIP similarity still yielded a high LPIPS value. This Supports that the generator is not reproducing near-duplicate training images. Representative $[\text{Generated} \mid \text{NN1--NN5}]$ grids with similarity annotations are shown in Fig.~\ref{fig:NN-Hajjv2-22} for direct inspection. Furthermore, we visualize the nearest neighbors in the worst case by deterministically selecting the generated samples with the smallest LPIPS distances to their closest training image retrieved using CLIP (NN1). CLIP is used for retrieval in embedding space (cosine similarity), and LPIPS is used to quantify perceptual similarity as illustrated in Fig.~\ref{fig:fig23}.

\par\vspace{\baselineskip}
\raggedbottom  
\noindent\begin{minipage}{\columnwidth}
\centering
\includegraphics[width=\linewidth,height=0.40\textheight]{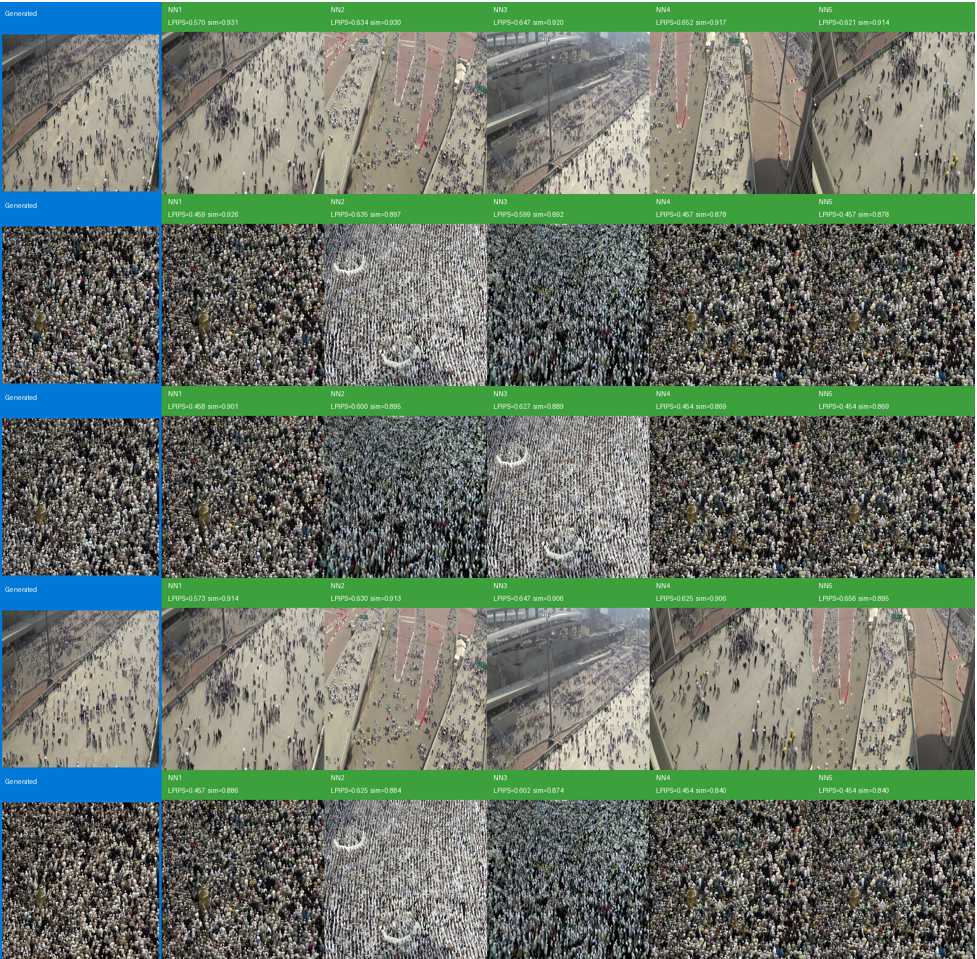}
\captionof{figure}{Worst-case nearest-neighbor grids. Each row shows a generated image followed by its top-5 nearest training images retrieved in CLIP embedding space (cosine similarity). NN1 denotes the closest neighbor; LPIPS is computed between the generated image and NN1.}
\label{fig:fig23}
\end{minipage}

\raggedbottom

\vspace{8pt}
Even under the selection of the top-30 most similar generated samples (minimum NN1 LPIPS), the closest generated samples yielded NN1 LPIPS values of 0.453 / 0.465 / 0.587 (min/median/max), which remained far from the low-distance regime (LPIPS: lower = more similar; higher = more different). In addition, visual inspection of the corresponding nearest-neighbor grids indicates that the generated samples remain perceptually distinct from their closest retrieved training frames.

For cross-domain confirmation, we repeated the same NN+LPIPS protocol using generated outputs on UCF\_CC\_50 (N = 200) and UCF-QNRF (N = 1,336) inputs. The NN1 LPIPS values remained high (0.671 ± 0.054 for UCF\_CC\_50; 0.702 ± 0.043 for UCF-QNRF). Furthermore, no low-distance mass indicative of near-duplicate reproduction was observed. Full cross-domain NN plots and grids are provided in Appendix A (Cross-Domain Plots and Grids).

\par\vspace{\baselineskip}

\noindent

\subsubsection{Samples Generated from P2P-H Model}
Fig.~\ref{fig:fig24} illustrates samples from the proposed model, highlighting its ability to reconstruct high-fidelity crowd images with plausible appearance variation under structurally consistent scene conditions. The visual quality of the synthesized images demonstrates successful adversarial training, in which the generator produces outputs that are visually consistent with real crowd appearances. These examples illustrate the capacity of the model to learn both local details and global crowd structure  under the proposed conditional synthesis framework.

\raggedbottom 
\par\vspace{\baselineskip}
\raggedbottom  
\noindent\begin{minipage}{\columnwidth}
\centering
\includegraphics[width=\linewidth,height=0.24\textheight]{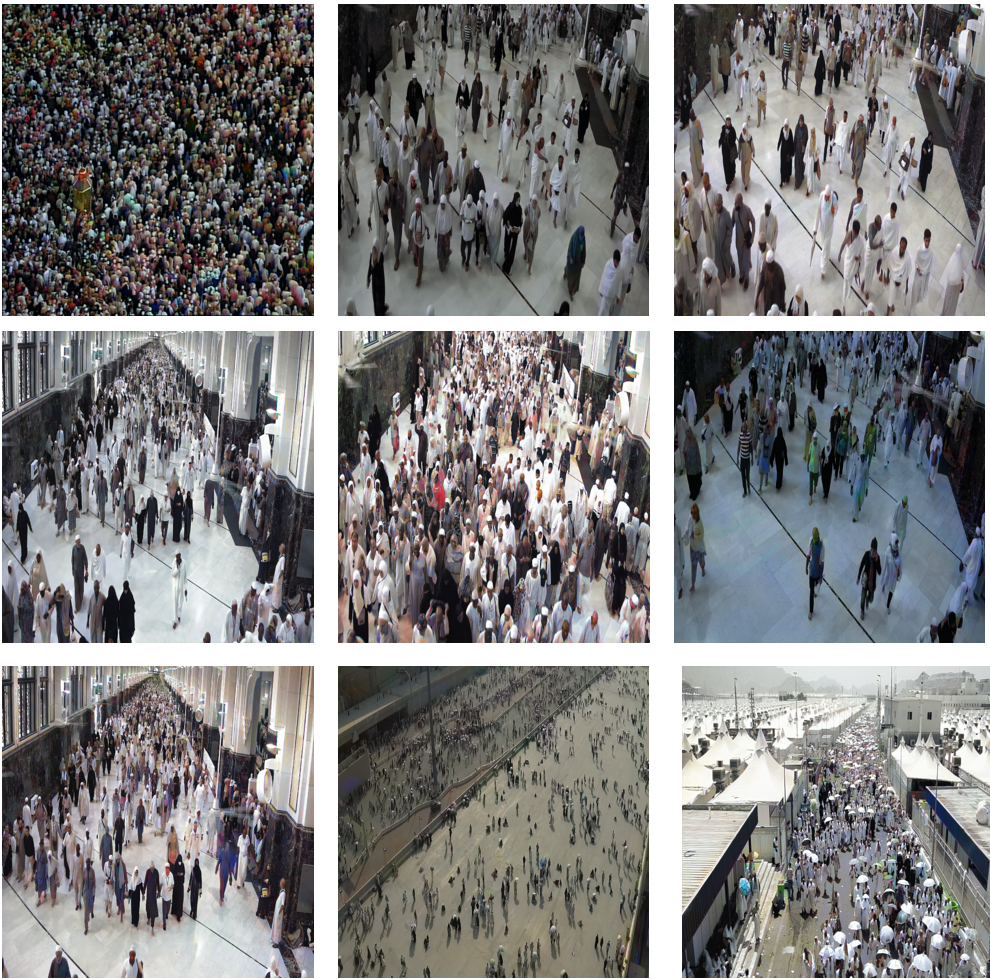}
\captionof{figure}{Sample outputs of the proposed P2P-H model, illustrating the visual fidelity of generated crowd scenes.}
\label{fig:fig24}
\end{minipage}

\FloatBarrier

\subsection{Crowd-Counting Performance with Real and Synthetic Training Data}
We report the experimental results on the CrowdH-Mix-469 dataset. The results in Table~\ref{tab:crowdhmix469_compare} show that augmenting the training set with selected P2P-H synthetic images reduced MAE for all five evaluated crowd-counting models. This suggests that the generated data provides useful training information for downstream Hajj crowd-counting under the evaluated protocol.
The largest improvement was observed for CSRNet, for which MAE decreased from 168.63 to 130.80, MSE decreased from 63,655.21 to 40,295.6, and RMSE decreased from 251.74 to 199.08. In addition, consistent gains were obtained for APGCC, with MAE decreasing from 135.88 to 121.23, MSE decreasing from 98,087.77 to 84,537.95, and RMSE from 287.61 to 268.50. Moreover, MCNN exhibited modest but consistent improvement across all three metrics.


\begin{table}[!htbp]
\caption{Comparison of crowd-counting performance on CrowdH-Mix-469 when training with real data only versus real and P2P-H synthetic data.}
\label{tab:crowdhmix469_compare}
\centering

\footnotesize
\setlength{\tabcolsep}{2pt}
\renewcommand{\arraystretch}{1.15}
\setlength{\arrayrulewidth}{0.4pt}

\begin{tabular*}{\columnwidth}{@{\extracolsep{\fill}}|c|c|c|c|c|c|c|}
\hline
\multicolumn{1}{|c|}{\multirow{2}{*}{Models}} &
\multicolumn{3}{c|}{Real data} &
\multicolumn{3}{c|}{Real + synthetic data} \\
\cline{2-7}
 & MAE & MSE & RMSE & MAE & MSE & RMSE \\
\hline
MCNN\cite{b3} & 387.87 & 445529.45 & 667.19 & 374.67 & 413610.6 & 642.93 \\
\hline
CSRNet\cite{b2} & 168.63 & 63655.21 & 251.74 & 130.80 & 40295.6 & 199.08 \\
\hline
DM-Count\cite{b66} & 226.78 & 246324.23 & 424.80 & 210.41 & 312343.61 & 512.95 \\
\hline
P2PNet\cite{b67} & 286.06 & 205717.45 & 452.22 & 261.56 & 218067.18 & 466.97 \\
\hline
APGCC\cite{b68} & 135.88 & 98087.77 & 287.61 & 121.23 & 84537.95 & 268.50 \\
\hline
\end{tabular*}
\end{table}
However, a more mixed pattern was observed for P2PNet and DM-Count. In both cases, MAE improved after adding synthetic data, whereas MSE and RMSE increased. This finding suggests that the synthetic augmentation helped reduce the average counting error, but did not consistently reduce larger-error cases across all architectural families. The downstream benefit of the generated data therefore appears to be architecture-dependent, with the strongest gains observed for convolutional density-regression models, particularly CSRNet.
These results provide direct downstream evidence that the proposed synthetic images are not only visually plausible but also practically useful for crowd-counting training. Nevertheless, the improvements are not uniform across all metrics or model architectures. Therefore, the present findings are interpreted as evidence of practical downstream utility under the evaluated setting rather than as proof of universal improvement across all crowd-counting architectures.

\section{Limitations}

Despite the strong performance of the P2P-H framework, it has several drawbacks:
\begin{itemize}
\item {Training data scale and coverage: Our model was trained using a relatively limited number of Hajj images. Therefore, expanding the dataset to thousands of diverse scenes would improve realism and cross-domain robustness.}

\item {Public-video sampling bias: Although the dataset includes multiple Hajj locations and density levels, using publicly available YouTube videos may introduce sampling bias related to camera-angle selection, video quality, and scene composition. As a result, CrowdH may not fully represent the visual distribution of operational deployment cameras. Future work should validate the framework using systematically sampled operational or surveillance footage.}
\item {Weak attribute supervision bias (edge, density proxy): Density labels are derived from image statistics (Canny edge density), which can be confounded by non-human edges (e.g., buildings, fences, textured floors, shadows). Consequently, density conditioning should be interpreted as a coarse semantic cue that may influence appearance tendencies rather than a guarantee of precise changes in the true number of individuals. Future work may integrate semantic segmentation or masking to improve background separation in density estimation.}
\item {The self-paired conditional objective constrains the structural diversity of the model. Since training pairs are self-derived, with the conditioning tensor deterministically computed from the same target image, P2P-H primarily learns attribute-conditioned reconstruction rather than synthesizing entirely novel scenes. This design stabilizes training and preserves geometry, but limits the ability to generate new crowd layouts when structural channels are fixed. This aspect impacts the effectiveness of density control when compared to unpaired or diffusion-based methods.}
	
\item {Computational requirements: Another weakness of P2P-H GANs is their computationally expensive nature for training and generation. Therefore, Google Colab Pro was used to train the GAN model.}
\item {Deployment constraints and safe-use concerns: Inference latency and computational budget are critical when deploying in a real world scenario, especially when generating high resolution images. The quality of the synthetic output should also be evaluated and validated before using it for model training or  operational analytics. In addition, synthetic images should never be utilized as "ground truth" for making safety-critical decisions without human evaluation and comparison with real data.}

\end{itemize}

\section{Conclusion and Future Work}

This study presented Pix2Pix-Hybrid (P2P-H), a structure-guided conditional synthesis framework for Hajj crowd-image augmentation under limited-data and privacy-sensitive conditions. The framework uses self-paired conditioning derived from edge, grayscale, weak density, and time-of-day cues to preserve the dominant crowd layout while rendering plausible Hajj crowd appearances. By integrating multi-channel conditioning, multi-scale adversarial discrimination, perceptual and feature-matching objectives, adaptive regularization, and EMA stabilization, P2P-H achieved improved conditional synthesis performance compared with the evaluated Pix2Pix and StyleGAN2-ADA baselines.

The study also introduced CrowdH, a synthetic Hajj crowd-image dataset, and CrowdH-Mix-469, an annotated mixed real-synthetic dataset for downstream crowd-counting evaluation. Experiments with five representative crowd-counting models showed that adding selected P2P-H synthetic images reduced MAE across all evaluated models, with the strongest improvement observed for CSRNet. However, improvements in MSE and RMSE were not uniform across all architectures, indicating that the benefit of synthetic augmentation is architecture-dependent rather than universal.

The proposed framework is designed for structure-guided conditional synthesis rather than fully unconstrained image generation. Because the conditioning tensor is self-derived, the model primarily supports appearance-level variation under fixed structural constraints. Future work will focus on expanding the dataset, integrating more diverse contextual inputs, and increasing structural diversity through alternative generative paradigms. In particular, cascaded pipelines in which structural crowd layouts are first synthesized and then rendered into photorealistic images may enable greater spatial diversity beyond the current self-paired conditioning scheme.

\FloatBarrier


\onecolumn

\appendices
\section{Cross-Domain Plots and Grids}

\begin{center}
\footnotesize
\setlength{\abovecaptionskip}{2pt}
\setlength{\belowcaptionskip}{2pt}

\begin{minipage}{0.47\textwidth}
\centering
\includegraphics[width=\linewidth,height=0.17\textheight,keepaspectratio]{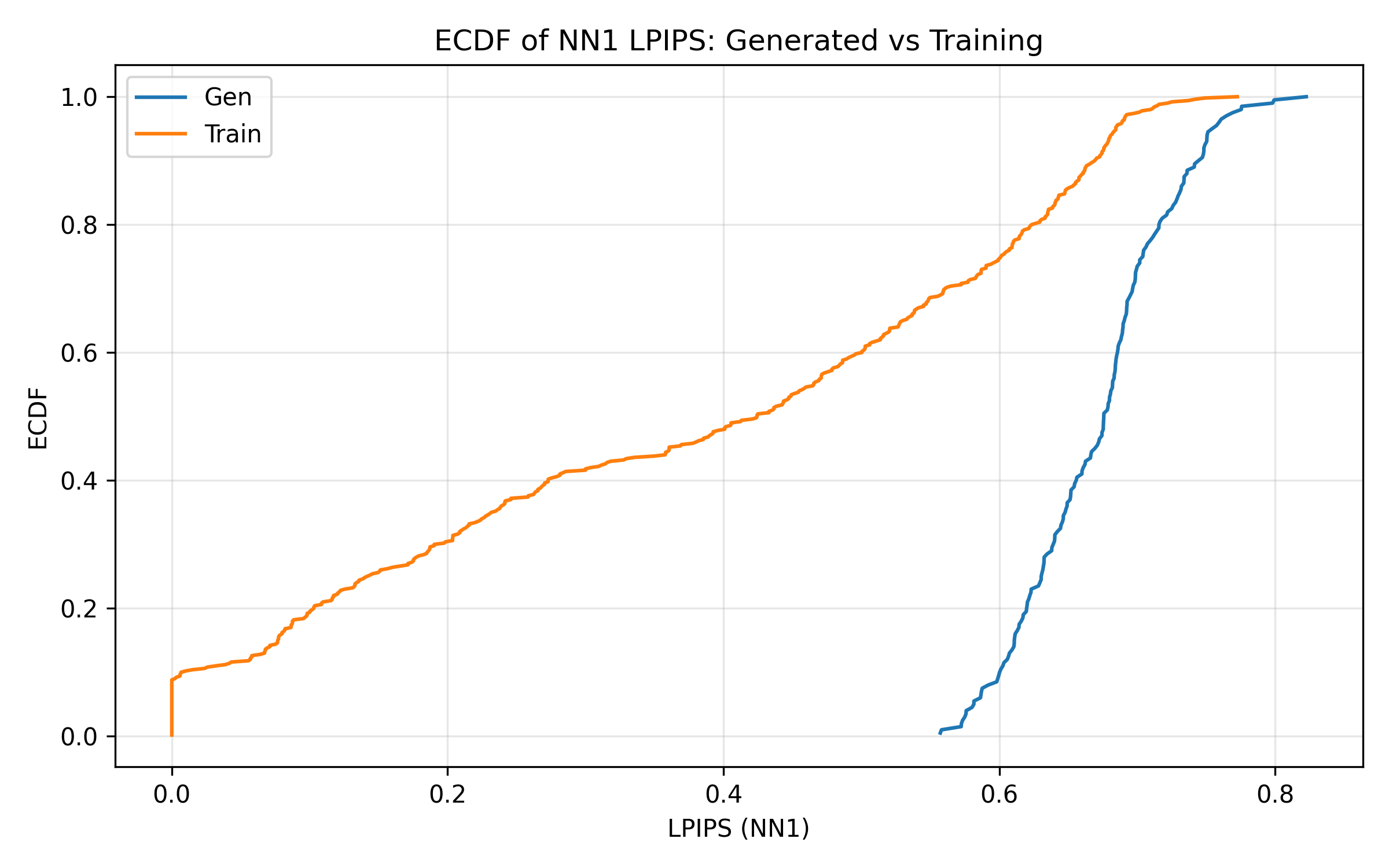}
\captionof{figure}{ECDF of LPIPS to the nearest neighbor (NN1) for UCF\_CC\_50.}
\label{fig:ECDF-ucf-cc-50}
\end{minipage}
\hfill
\begin{minipage}{0.47\textwidth}
\centering
\includegraphics[width=\linewidth,height=0.17\textheight,keepaspectratio]{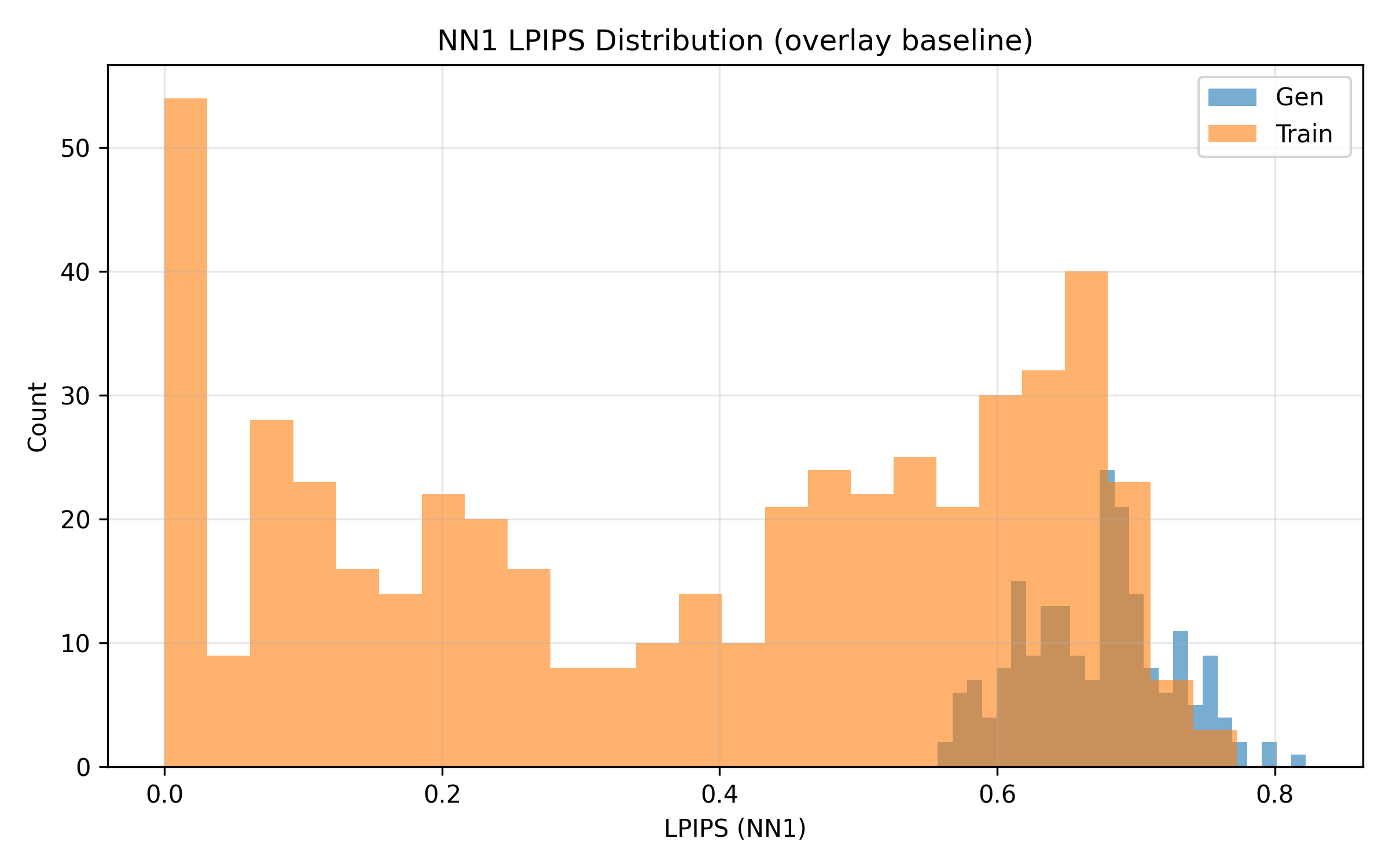}
\captionof{figure}{Distribution overlay showing generated distances compared to training baseline for UCF\_CC\_50.}
\label{fig:ucf-cc-50-histogram}
\end{minipage}

\vspace{6pt}

\begin{minipage}{0.47\textwidth}
\centering
\includegraphics[width=\linewidth,height=0.17\textheight,keepaspectratio]{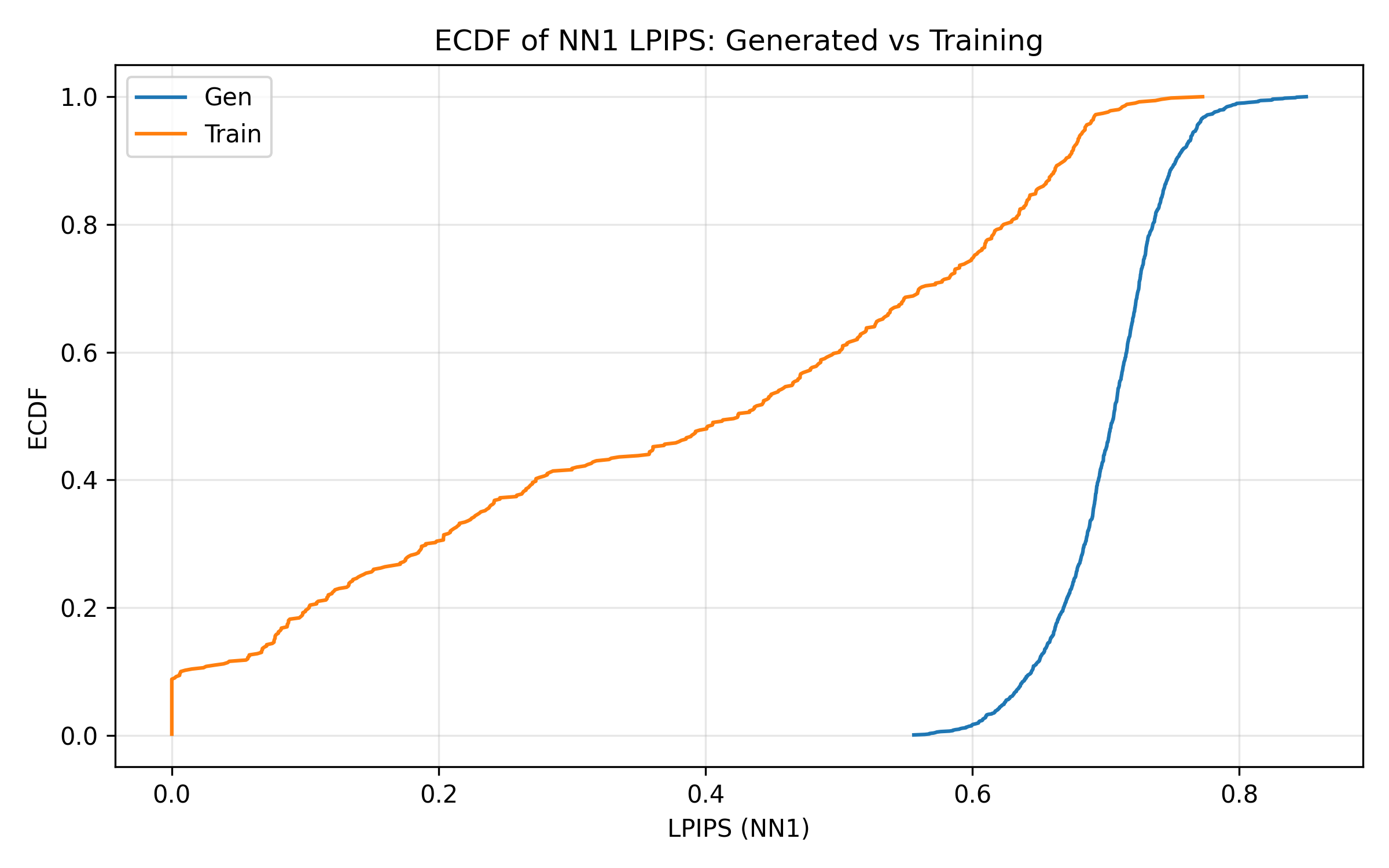}
\captionof{figure}{ECDF of LPIPS to the nearest neighbor (NN1) for UCF-QNRF.}
\label{fig:ECDF-UCF_QNRF}
\end{minipage}
\hfill
\begin{minipage}{0.47\textwidth}
\centering
\includegraphics[width=\linewidth,height=0.17\textheight,keepaspectratio]{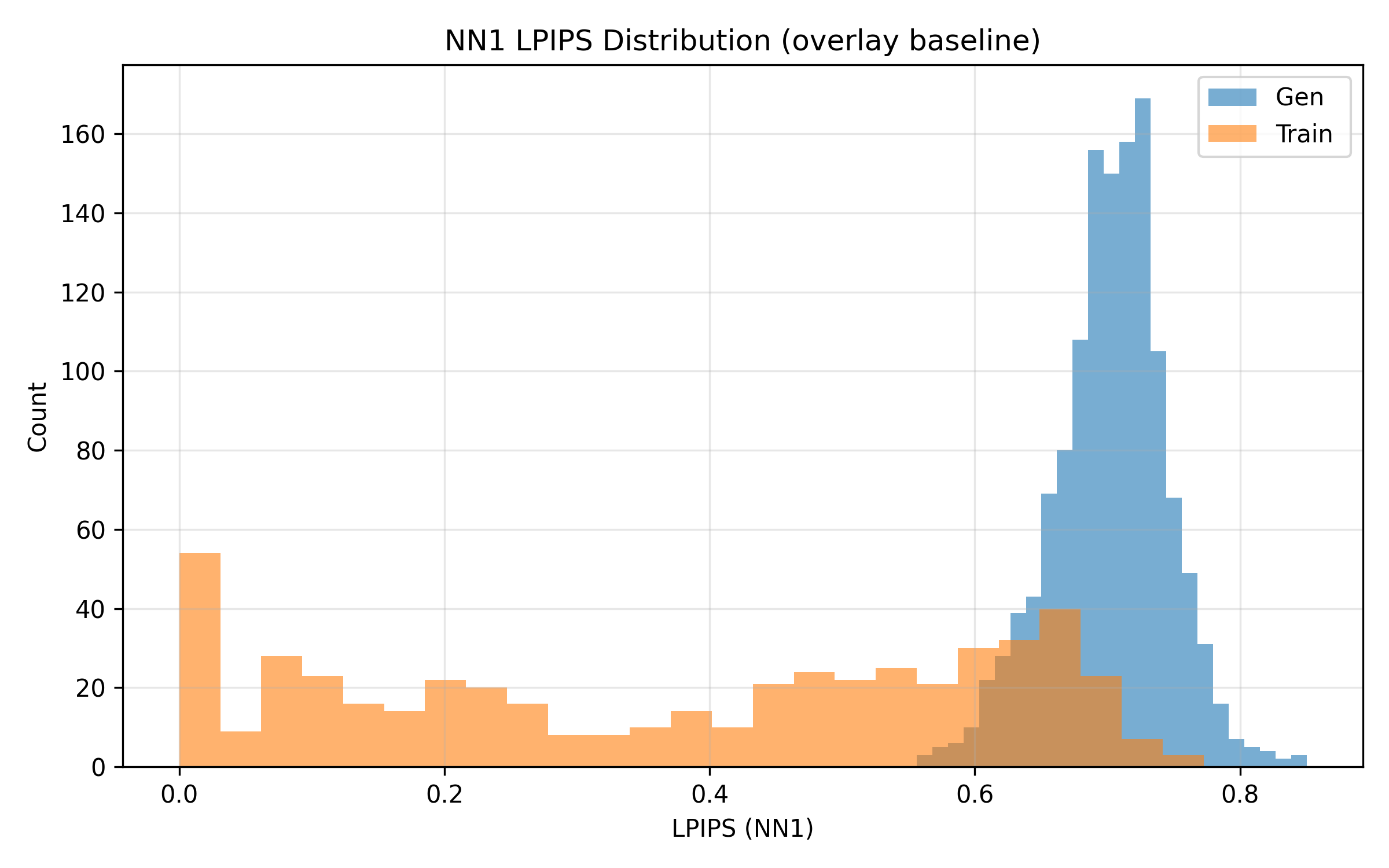}
\captionof{figure}{Distribution overlay showing generated distances compared to training baseline for UCF-QNRF.}
\label{fig:UCF-QNRF-histogram}
\end{minipage}

\vspace{8pt}

\captionof{table}{NN1 LPIPS statistics across in-domain and cross-domain inputs; in all cases, 0\% of generated samples fall below LPIPS $<0.2$ against the training data.}
\label{tab:nn_lpips_summary}

\scriptsize
\setlength{\tabcolsep}{2.3pt}
\renewcommand{\arraystretch}{1.08}
\setlength{\arrayrulewidth}{0.4pt}

\begin{tabularx}{0.98\textwidth}{|
>{\raggedright\arraybackslash}X |
>{\raggedright\arraybackslash}p{2.5cm} |
>{\centering\arraybackslash}p{1.0cm} |
>{\raggedright\arraybackslash}p{2.4cm} |
>{\centering\arraybackslash}p{0.85cm} |
>{\centering\arraybackslash}p{0.95cm} |
>{\centering\arraybackslash}p{0.85cm} |
>{\raggedright\arraybackslash}p{2.5cm} |}
\hline
Generated input dataset &
Retrieval pool (training) &
n\_gen &
NN1 LPIPS (mean $\pm$ std) &
min & median & max &
NN1 CLIP sim (mean / max) \\
\hline
HAJJv2 (in-domain) &
\multirow{3}{*}{train ($N{=}993$)} &
2{,}039 & 0.6474 $\pm$ 0.0418 & 0.4529 & 0.6456 & 0.8246 & 0.8435 / 0.9367 \\
\cline{1-1}\cline{3-8}
UCF\_CC\_50 (cross-domain) &
& 200 & 0.6711 $\pm$ 0.0535 & 0.5572 & 0.6757 & 0.8225 & 0.7644 / 0.9522 \\
\cline{1-1}\cline{3-8}
UCF-QNRF (cross-domain) &
& 1{,}336 & 0.7021 $\pm$ 0.0432 & 0.5561 & 0.7059 & 0.8502 & 0.7445 / 0.9559 \\
\hline
\end{tabularx}

\end{center}
\FloatBarrier

\begin{figure*}[!htbp]
\centering
\includegraphics[width=\textwidth,height=0.80\textheight,keepaspectratio]{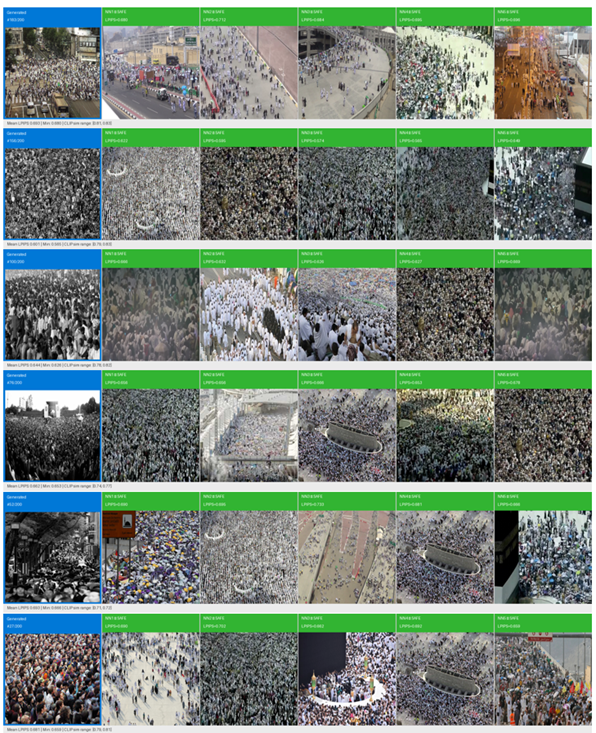}
\caption{Samples output of our proposed P2P-H illustrating the visual fidelity of generated crowd scenes from the UCF\_CC\_50 dataset.}
\label{fig:grid-ucf-cc50}
\end{figure*}
\FloatBarrier
\FloatBarrier
\clearpage
\begin{figure*}[!t]
\centering
\includegraphics[width=\textwidth,height=0.80\textheight,keepaspectratio]{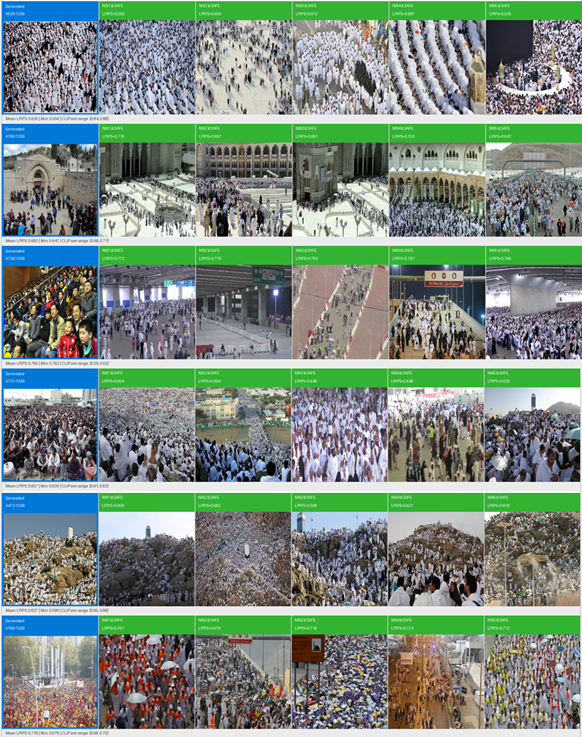}
\vspace{-2mm} 
\caption{Samples output of our proposed P2P-H illustrating the visual fidelity of generated crowd scenes from the UCF-QNRF dataset.}
\label{fig:final_grid_ucfqnrf}
\end{figure*}
\vspace{-2mm} 

\FloatBarrier

\FloatBarrier
\twocolumn

\FloatBarrier

\end{document}